\documentclass[final,hidelinks,onefignum,onetabnum,nohypdvips]{siamart251216}
\usepackage{graphicx}
\usepackage{amssymb}
\usepackage{float}
\usepackage{algorithmic}
\usepackage{multirow}
\usepackage{booktabs}

\newtheorem*{Van Dyke}{Van Dyke Matching Principle}
\usepackage{subcaption}
\usepackage{makecell}
\newsiamremark{remark}{Remark}

\headers{PVD-ONet}{T. Sun and J. Zu}

\title{PVD-ONet: A Multi-scale Neural Operator
Method for Singularly Perturbed Boundary Layer Problems%
\thanks{\textbf{Funding:}{ The work of J. Zu was supported by the Science and Technology Development Plan Project of Jilin Province under Grant No.~20250102016JC.}}}

\author{
Tiantian Sun\thanks{School of Mathematics and Statistics, Center for Mathematics and Interdisciplinary Sciences, Northeast Normal University, Changchun 130024, P.R. China}
\and
Jian Zu\footnotemark[1]\thanks{Corresponding author (\email{zuj100@nenu.edu.cn}).}
}

\ifpdf
\hypersetup{
  pdftitle={PVD-ONet: A Multi-scale Neural Operator Method for Singularly Perturbed Boundary Layer Problems},
  pdfauthor={Tiantian Sun and Jian Zu}
}
\fi

\begin{document}

\maketitle

\begin{abstract}
Physics-informed neural networks and physics-informed DeepONet excel at solving partial differential equations but often struggle with singularly perturbed problems. To address this limitation, we propose two novel data-free frameworks: the \emph{Prandtl--Van Dyke neural network} (PVD-Net) and its operator-learning extension, the \emph{Prandtl--Van Dyke Deep Operator Network} (PVD-ONet). Each framework provides leading-order and high-order versions to meet different modeling requirements. The leading-order PVD-Net integrates Prandtl's matching condition for improved stability, while the high-order PVD-Net employs Van Dyke's matching principle to capture fine-scale boundary layer structures with higher accuracy. Extending PVD-Net to operator learning, PVD-ONet enables efficient prediction for families of boundary layer problems without retraining. Numerical experiments on representative one- and two-dimensional problems demonstrate that the proposed methods outperform existing baselines. Beyond forward prediction, the framework also supports inverse problems by inferring the scaling exponent governing boundary layer thickness from sparse observations, highlighting its potential for practical applications.
\end{abstract}



\section{Introduction}
Singular perturbation problems are prevalent in various scientific and engineering disciplines, including fluid mechanics \cite{white2006viscous,white2017fluid}, aerodynamics \cite{schoppa1998large,anderson1989hypersonic,anderson2011ebook}, solid mechanics \cite{reiss1962symmetric,chien1947large}, and biological transport \cite{arzani2016lagrangian}. Such problems typically involve a small parameter multiplying the highest-order derivative \cite{nayfeh2024perturbation}, causing rapid variations in narrow regions and slow variations in the remaining domain, leading to multiscale behavior and boundary layers. In 1904, Prandtl \cite{prandtl1904fluid} proposed boundary layer theory, which effectively addressed singular perturbation difficulties and established the foundation for aerodynamic drag and lift calculations. In 1921, von Kármán \cite{von1921uber} derived the boundary layer momentum integral equation, further advancing engineering applications. Van Dyke \cite{van1962higher-1,van1962higher-2,van1964higher} later developed high-order matching methods for boundary layer problems, providing a foundation for accurate high-order approximations. O'Malley \cite{o1968boundary,o1968topics,o1971boundary} and Nayfeh \cite{nayfeh2024perturbation,nayfeh1964generalized,nayfeh1965perturbation} further established rigorous theories for singular perturbation and boundary layer analysis.
To solve these problems, traditional mesh-based methods, including the Finite Element Method (FEM) \cite{hughes2012finite}, Finite Volume Method (FVM) \cite{eymard2000finite}, and Finite Difference Method (FDM) \cite{strikwerda2004finite}, are widely used. However, they often require fine meshes to resolve multiscale structures and provide only numerical approximations without explicit analytical structure. In contrast, singular perturbation methods are mesh-free and provide near-analytical solutions by directly capturing the intrinsic behavior of the problems. Therefore, they have attracted sustained attention in science and engineering.

In recent years, the rapid advancement of artificial intelligence has sparked growing interest in deep learning methods. Scientific Machine Learning (SciML) has emerged as an alternative to traditional numerical methods for complex system modeling \cite{ling2016reynolds,srinivasan2019predictions,jiang2021interpretable}. Physics-Informed Neural Networks (PINNs) \cite{raissi2019physics,lu2021deepxde} incorporate governing equations as soft constraints in the loss function, enabling neural networks to satisfy physical laws while reducing reliance on labeled data. PINNs have been successfully applied to various fields, including biology \cite{arzani2021uncovering}, chemical kinetics \cite{ji2021stiff} and fluid mechanics \cite{cai2021physics,mao2020physics}.
However, PINNs often encounter convergence difficulties for multi-scale problems \cite{karniadakis2021physics}. Existing studies \cite{arzani2023theory} indicate that this issue persists despite advanced techniques such as domain decomposition \cite{jagtap2020extended} and importance sampling \cite{nabian2021efficient}. This challenge may arise from the difficulty of a single PINN in simultaneously capturing variations across different scales \cite{wang2021eigenvector}. From the perspective of approximating analytical solutions, perturbation techniques exhibit a natural connection with deep learning methods, offering a promising pathway to address the aforementioned challenges.
Since the method of matched asymptotic expansions is a classical approach for handling multi-scale problems in singular perturbation theory \cite{nayfeh2024perturbation,van1962higher-1,van1962higher-2,van1964higher, kaplun1967fluid}, a natural idea is to combine it with the PINNs framework. Notably, unlike traditional perturbation approaches requiring intricate derivations, combining matched asymptotic expansions with PINNs eliminates the need for manual analysis while retaining the advantage of operating directly on the governing equations under perturbation theory.

Recent works have started exploring hybrid approaches that integrate matched asymptotic expansions with  PINNs to better model singularly perturbed problems. Arzani et al. proposed Boundary-layer PINNs (BL-PINNs) \cite{arzani2023theory}, which employ two separate neural networks to approximate the inner and outer solutions. These solutions are connected through Prandtl’s matching principle to obtain a leading-order approximation. However, BL-PINNs do not construct a composite solution over the entire domain, and their piecewise formulation may therefore lead to reduced accuracy in the transition region. Zhang and He \cite{zhang2024multi} proposed the Multi-Scale-Matching Neural Networks (MSM-NN) framework to address problems with multiple boundary layers. By introducing exponential stretched variables  in the boundary layers, their method avoids semi-infinite domain issues and improves solution accuracy. Other works \cite{cao2023physics,huang2024multi} have combined asymptotic expansions with PINNs, but they primarily focus on leading-order approximations and struggle to capture higher-order effects, limiting their ability to achieve high-accuracy solutions.
{In addition, in these approaches, the stretched variables are constructed using fixed scaling exponents that are assumed to be known a priori, which restricts their flexibility in practical scenarios.}
Meanwhile, the application of higher-order matching techniques, such as Van Dyke matching \cite{van1962higher-1,van1962higher-2,van1964higher}, to multi-scale problems remains largely unexplored. 

In this paper, our first contribution is the proposal of the Prandtl-Van Dyke neural network (PVD-Net), a novel deep learning framework for solving boundary layer problems, which consists of a leading-order PVD-Net and a high-order PVD-Net. For leading-order PVD-Net, two neural networks are used to learn
the inner and outer solutions separately. By constructing a global composite solution through the Prandtl matching principle, we obtain a uniformly valid approximation—this is the crucial feature that significantly enhances numerical accuracy of BL-PINNs in the transition region. For high-order PVD-Net, we propose a novel five-network architecture incorporating Van Dyke's matching principle. This framework systematically captures finer-scale variations through higher-order asymptotic matching, enabling significantly improved numerical accuracy compared to leading-order PVD-Net.  {Moreover, the proposed framework can be extended to inverse problems, enabling the identification of key scaling exponents and significantly improving its applicability in practical scenarios.} 

The second contribution of our work is the proposal of the \emph{Prandtl--Van Dyke Deep Operator Network} (PVD-ONet), an operator learning framework for a family of boundary layer problems. Inspired by the core idea of PVD-Net, PVD-ONet consists of two components: the leading-order PVD-ONet and the high-order PVD-ONet. Although PVD-Net enables efficient solutions for boundary layer problems under fixed settings, it requires retraining when system parameters (e.g.,boundary conditions) change. To overcome this limitation, we extend the framework to operator learning methods \cite{li2020neural,lu2021learning,li2020fourier,kovachki2023neural}. Unlike PINNs, neural operators approximate solution operators parametrically, learning entire PDE families in a single training process and allowing fast online inference for new instances.
Advanced neural operator architectures, such as the Fourier Neural Operator (FNO) \cite{li2020fourier} and Deep Operator Network (DeepONet) \cite{lu2021learning}, have been developed for this purpose. Li et al. \cite{du2023approximation} applied DeepONet to learn families of boundary layer problems, allowing adaptation to different initial conditions without retraining. However, despite the impressive performance of these methods, they are predominantly used in a data-driven manner. Typically, due to the extremely thin structure of boundary layers, obtaining high-fidelity measurement data is exceedingly difficult and often infeasible in practice. Physics-informed DeepONet (PI-DeepONet) \cite{wang2021learning} reduces this dependence by incorporating physical constraints, but it still encounters convergence difficulties for multi-scale problems. Therefore, learning families of boundary layer problems solely from governing equations remains largely unexplored in current research.

Thus, our proposed PVD-ONet fills this gap by extending the core idea of PVD-Net to the operator learning framework.  The leading-order PVD-ONet inherits the two-network architecture of PVD-Net, using two DeepONet models to approximate the inner and outer solutions, which are connected through Prandtl’s matching conditions and combined into a composite solution over the entire domain. The high-order PVD-ONet extends this framework to five DeepONet models for higher-accuracy approximation and employs Van Dyke’s matching principle to connect solutions across different regions.   A detailed comparison of neural network-based methods for boundary layer approximation is provided (see Table \ref{tab:method_comparison}). The contributions of this work can be summarized as follows:
\begin{itemize}
    \item We propose PVD-Net for multiscale problems. Its leading-order version enhances BL-PINNs with a composite solution in transition regions, while its high-order version combines multi-network architecture with Van Dyke matching to resolve fine-scale structures. PVD-Net incorporates singular perturbation theory into the model design and is trained solely from governing equations, eliminating the need for high-fidelity training data.

    \item We propose PVD-ONet, which extends PVD-Net to operator learning and predicts solutions for new inputs in a single forward pass, enabling efficient solution of boundary layer problem families without retraining.

    \item Beyond forward prediction, our framework addresses inverse problems by inferring asymptotic scaling parameters from sparse observations rather than prescribing them a priori.

    \item Comparisons on representative one- and two-dimensional problems show that the leading-order models outperform existing baselines and the high-order models further improve accuracy.
\end{itemize}

\begin{table}[htbp]
\centering
\footnotesize
\caption{Comparison of neural network methods for singular perturbation problems.}
\label{tab:method_comparison}
\resizebox{\linewidth}{!}{%
\begin{tabular}{cccccc}
\toprule
\textbf{Method} & \textbf{Training} & \textbf{New Instance} & \textbf{Leading-Order} & \textbf{High-Order} & \textbf{Smooth} \\
       & \textbf{Speed}    
       & \textbf{Inference}     
       & \textbf{Approximation} 
       & \textbf{Approximation} 
       & \textbf{Solution}  \\
\midrule
BL-PINNs        & Fast  & No  & Yes & No  & No  \\
MSM-NN          & Fast  & No  & Yes & No  & Yes \\
\textbf{PVD-Net (Ours)}  & Fast  & No  & Yes (Stable, Simple) & Yes (High Precision) & Yes \\
\textbf{PVD-ONet (Ours)} & Slow  & Yes & Yes (Stable, Simple) & Yes (High Precision) & Yes \\
\bottomrule
\end{tabular}%
}
\end{table}
The remainder of this paper is organized as follows. Section~\ref{problem setup} introduces the singular perturbation problems and matched asymptotic expansions. Section~\ref{methods} presents the PVD-Net and PVD-ONet frameworks, and Section~\ref{numerical experiments} provides numerical experiments. Section~\ref{conclusion} concludes the paper. The supplementary material contains additional theoretical and computational details, including the proof of Theorem~1, the variable-coefficient formulation and results, data-driven learning of the scaling exponent $\lambda$, and the implementation and mathematical details of PVD-Net. 


\section{Problem Setup and Matched Asymptotic Expansion Techniques}\label{problem setup}
\subsection{Problem Statement}
Perturbation methods are commonly used to analyze differential equations containing a small parameter $\varepsilon \ll 1$. It can generally be classified into two types: regular and singular perturbations. In regular perturbation problems, the solution varies smoothly with respect to $\varepsilon$, and setting $\varepsilon = 0$ yields a reduced problem whose solution remains a good approximation throughout the entire domain. However, many problems of practical importance fall into the category of singular perturbation problems, where the small parameter $\varepsilon$ multiplies the highest-order derivative in the differential equation. These problems are particularly challenging because the solution behavior for $\varepsilon = 0$ differs fundamentally from the behavior as $\varepsilon \to 0$. This discrepancy gives rise to sharp gradients or boundary layers, making the construction of accurate approximations considerably more difficult than in regular perturbation problems. 

Consider a differential equation of the following form:
\begin{align}\label{Equation}
    \mathcal{F}(u,x,\varepsilon)=0,\quad x \in \Omega \subseteq \mathbb{R}^{n},
\end{align}
which subjects to suitable boundary conditions,
\begin{align}
    \mathcal{B}(u,\varepsilon)=0\quad on \quad \partial\Omega,
\end{align}
where $\varepsilon\ll1$ is the small parameter,
$u:\Omega \to \mathbb{R}$ is the unknown solution of equation (\ref{Equation}), $\mathcal{B}(u,\varepsilon)$ represents boundary conditions such as Dirichlet, Neumann, and so on, depending on the specific problem.  We consider an asymptotic expansion of the unknown solution $u(x;\varepsilon)$ in powers of the small parameter $\varepsilon$:
\begin{align*}
u(x;\varepsilon)=u_0(x)+\varepsilon u_1(x)+\varepsilon^2u_2(x)+\cdots .
\end{align*}
In regular perturbation problems, substituting this expansion into the original equation (\ref{Equation}) and matching terms at each order of 
$\varepsilon$ typically yields a sequence of well-posed problems for $u_0,u_1,u_2\dots$, valid uniformly across the entire domain. 

However, when the small parameter $\varepsilon$ multiplies the highest-order derivative term, the regular perturbation method breaks down, and the problem falls into the category of singular perturbation problems, which are notably more challenging due to the presence of boundary layers and the failure of uniform asymptotic expansions. In such cases, directly setting $\varepsilon=0$ typically reduces the order of the differential equation, which results in redundant boundary conditions and leads to an ill-posed problem. A prominent feature of this type of problem is the appearance of boundary layers, where the solution exhibits abrupt changes in certain local regions, while in the remaining regions, the solution changes relatively smoothly. The thickness of the boundary layer is usually related to $\varepsilon$. As $\varepsilon\to 0$, the boundary layer becomes extremely narrow. This phenomenon makes it difficult for conventional regular asymptotic expansion methods to effectively describe the global behavior of the solution. To overcome this difficulty, multi-scale analysis needs to be introduced, particularly the method of matched asymptotic expansion. 
\subsection{Matched Asymptotic Expansion}
For region $\Omega_o\subset \Omega$ away from the boundary layer, the outer expansion of the solution is defined as:
\begin{align*}
    u^{o}(x;\varepsilon)=\sum_{k=0}^{\infty} \varepsilon^k \mathbf{\varphi }_{k}(x),
\end{align*}
where $\varphi_k:\Omega_o\to\mathbb{R}$ represents the asymptotic solutions for each scale in the region $\Omega_o$. By substituting the outer expansion 
$u^{o}(x;\varepsilon)$ into (\ref{Equation}),  the governing equation becomes
\begin{align*}
\mathcal{F}_0(\varphi_0,x)+\varepsilon\mathcal{F}_1(\varphi_0,\varphi_1,x)+\varepsilon^2\mathcal{F}_2(\varphi_0,\varphi_1,\varphi_2,x)+\dots=0,
\end{align*}
where $\mathcal{F}_k$ ($k=0,1,2,\dots$) are differential operators arising from the expansion of $\mathcal{F}$. According to the theory of singular perturbations, the outer solution satisfies only the boundary conditions on the outer boundary and ignores the conditions near the boundary layer, i.e.,
\begin{align*}
\mathcal{B}^o(u^o,\varepsilon)=0.
\end{align*}

\begin{figure}
    \centering
    \includegraphics[width=0.4\linewidth]{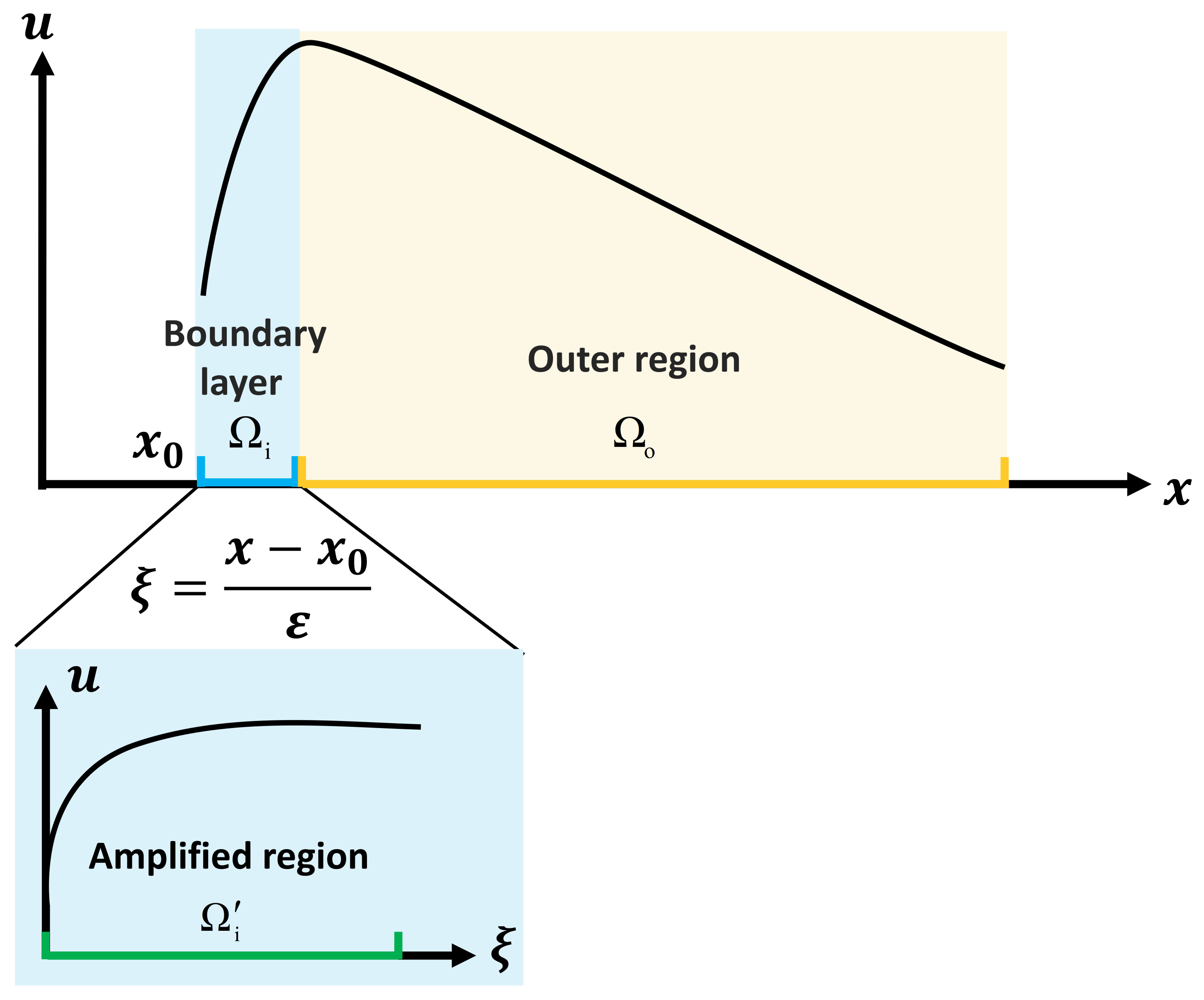}
    \caption{Schematic diagram of the boundary layer scale transformation. The diagram illustrates the evolution of the boundary layer's scale. The domain is decomposed into two parts: a boundary layer region $\Omega_i$ and an outer region $\Omega_o$. By introducing the transformation $\xi = \frac{x - x_0}{\varepsilon}$, the boundary layer region $\Omega_i$ is mapped to an amplified region $\Omega_i^\prime$ in the stretched coordinate.}
    \label{expansion}
\end{figure}

For the boundary layer region $\Omega_i$, a new stretched variable $\xi =\frac{x-x_0}{\varepsilon^\lambda}\in \Omega_i^{\prime}$ needs to be introduced to amplify the boundary layer, where $x_0$ is the location of the boundary layer.  In this paper, for convenience, we set the stretching exponent $\lambda=1$, and the choice of scaling will be discussed in another paper. A schematic diagram of the boundary layer scale transformation is shown in Figure \ref{expansion}. The rescaling of the original equation (\ref{Equation})  leads to a new form of the equation in terms of $\xi$. Specifically, after applying the stretching transformation, the governing equation becomes
\begin{align}\label{Equation-in}
    \tilde{\mathcal{F}}(u,\xi,\varepsilon)=0,
\end{align}
where $\tilde{\mathcal{F}}$ represents the transformed operator, with derivatives taken with respect to $\xi$ instead of $x$. The new equation describes the behavior of the solution in the boundary layer region. 
Next, the solution in the boundary layer can be expressed as an asymptotic expansion in $\varepsilon$ as follows:
\begin{align*}
    u^{i}(\xi ;\varepsilon)=\sum_{k=0}^\infty\varepsilon^k\psi_k(\xi ),
\end{align*}
where $\psi_k:\Omega^{\prime}_i\to\mathbb{R}$ are coefficient functions that describe the behavior of the solution in the boundary layer region. Substituting this expansion into the rescaled equation (\ref{Equation-in}), we obtain the following asymptotic expansion of the operator $\tilde{\mathcal{F}}$:
\begin{align*}
\tilde{\mathcal{F}}_0(\psi_0,\xi)+\varepsilon\tilde{\mathcal{F}}_1(\psi_0,\psi_1,\xi)+\varepsilon^2\tilde{\mathcal{F}}_2(\psi_0,\psi_1,\psi_2,\xi)+\dots=0,
\end{align*}
where $\tilde{\mathcal{F}}_k$ ($k=0,1,2,\dots$) are the differential operators that arise from the expansion of $\tilde{\mathcal{F}}$. The inner solution satisfies only the boundary conditions within the boundary layer region, i.e.,
\begin{align*}
\mathcal{B}^i(u^i,\varepsilon)=0.
\end{align*}

Asymptotic solutions in different regions are connected via matching techniques to ensure a uniformly valid approximation over the entire domain. Prandtl’s matching principle \cite{prandtl1904fluid} effectively matches leading-order inner and outer solutions. For higher-order approximations in singular perturbation problems, the Van Dyke matching method \cite{van1962higher-1,van1962higher-2,van1964higher} is used, allowing more accurate composite solutions by ensuring consistency in higher-order expansions. Specifically, the principles are as follows:
\paragraph{Prandtl Matching Principle \cite{nayfeh2024perturbation}} The outer limit of the inner solution is equal to the inner limit of the outer solution. 

\paragraph{Van Dyke Matching Principle \cite{nayfeh2024perturbation}} 
The $m$-term inner expansion of the $n$-term outer expansion is equal to the $n$-term outer expansion of the $m$-term inner expansion, where $m$, $n$ are arbitrary integers. 

Finally, we can construct a single solution that is uniformly valid across the entire domain, known as the composite solution \cite{nayfeh2024perturbation, erdelyi1961expansion}, i.e.,
\begin{align*}
    u^c(x)= u^{i}\left(\frac{x-x_0}{\varepsilon}\right)+u^{o}(x)-u^{o,i},\quad x\in\Omega,
\end{align*}
where $u^{o,i}$ denotes the Prandtl matching term or the Van Dyke matching term. 
The introduction of the composite solution significantly improves the accuracy in the transition region between the inner and outer layers, mitigating the loss of precision.  

\begin{remark}
    Although the inner solution $u^i\left(\frac{x-x_0}{\varepsilon}\right)$ is formally defined in the inner region $\Omega_i$, it decays rapidly to the matching term $u^{o,i}$ outside this region and thus can be smoothly extended to the entire domain $\Omega$. Similarly, the outer solution $u^o(x)$ is originally defined in the outer region $\Omega_o$, but it can be smoothly extrapolated into the inner region. The matching term $u^{o,i}$
  serves to remove the overlapping contribution, ensuring that the composite solution remains uniformly valid across the whole domain $\Omega$.
\end{remark}


\section{Methods}\label{methods}
\subsection{Prandtl-Van Dyke Neural Network (PVD-Net)}
\begin{figure}[h]
    \centering
    \includegraphics[width=0.85\linewidth]{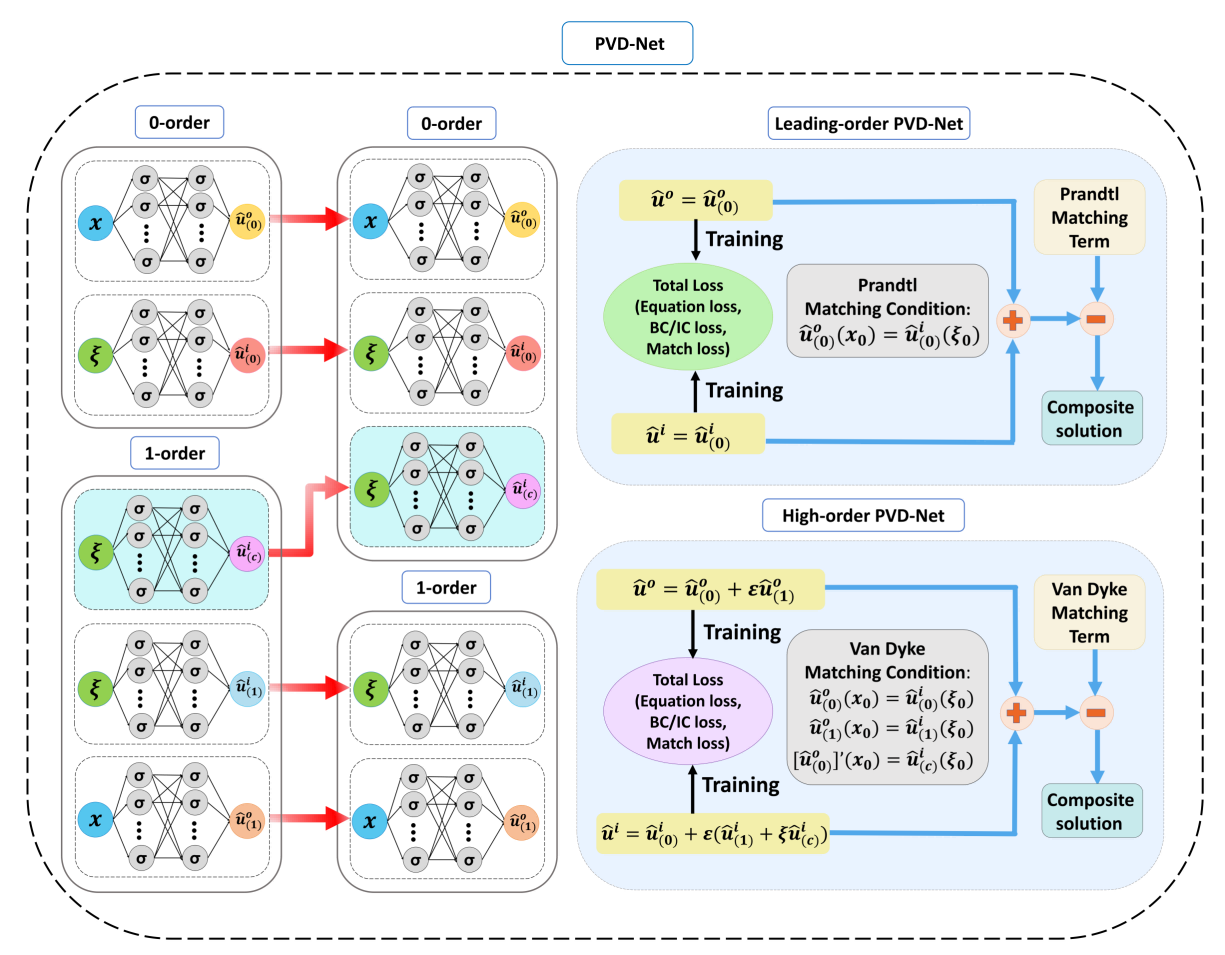}
    \caption{Network architecture of PVD-Net. The leading-order version uses outer and inner networks with Prandtl matching. The high-order version employs five networks to construct leading- and first-order approximations through Van Dyke matching, with one of the first-order correction networks naturally reducing to the leading-order formulation. Both versions use composite solutions as the final output to ensure smooth transitions.}
    \label{PVD-Net-architecture}
\end{figure}

In this section, we introduce the PVD-Net framework, including its leading-order and high-order formulations. The former is based on Prandtl matching, while the latter incorporates Van Dyke matching to retain higher-order asymptotic corrections. The architecture of the PVD-Net is illustrated in Figure \ref{PVD-Net-architecture}.

\vspace{-20pt}

\subsubsection{Leading-order PVD-Net}\label{Leading-order PVD-Net}
For the leading-order PVD-Net, we employ a fully connected neural network $\hat{u}^{o}_{(0)}(x;\theta_1)$ as a surrogate model for the outer solution of (\ref{Equation}). Similarly, another fully connected neural network $\hat{u}^{i}_{(0)}(\xi; \theta_2)$ is used to represent the inner solution. Here $x$ is the outer variable, $\xi =\frac{x-x_0}{\varepsilon}$ is the stretched inner variable, $\theta_1$, $\theta_2$ are the trainable parameter. We simplify notation by writing $\hat{u}^{o}_{(0)}(x)$ and $\hat{u}^{i}_{(0)}(\xi)$ for $\hat{u}^{o}_{(0)}(x;\theta_1)$ and $\hat{u}^{i}_{(0)}(\xi;\theta_2)$, respectively, whenever no confusion arises. The Prandtl matching principle can be denoted by
\begin{align*}
    \lim_{x\to x_0}\hat{u}^{o}_{(0)}(x)=\lim_{\xi \to\infty}\hat{u}^{i}_{(0)}(\xi).
\end{align*}
As computers are incapable
of handling infinity, we approximate it by choosing a sufficiently large but finite value $\xi_0$.  This is justified by the fact that the inner solution decays exponentially in the outer region, so choosing a sufficiently large $\xi_0$ yields the same effect as taking $\xi\to\infty$.

Let $\mathcal{T} = \mathcal{T}_o \cup \mathcal{T}_i \cup \mathcal{T}_m \cup \mathcal{T}_b$, where $\mathcal{T}_o \subset \Omega_o$ and $\mathcal{T}_i \subset \Omega_i'$ are finite sets of residual points sampled from their respective domains, respectively. The set $\mathcal{T}_b=\Gamma^i\cup\Gamma^o$
consists of boundary residual points,
with $\Gamma^i:=\partial \Omega\setminus\partial \Omega_o$, $\Gamma^o:=\partial \Omega\setminus\partial\Omega_i$. $\mathcal{T}_m = \{x_0, \xi_0\}$ denotes the matching points under the matching principle. 
The neural networks are optimized by the following loss function:
\begin{align}\label{loss-Leading-PVD-Net}
    \mathcal{L}(\theta;\mathcal{T}):=\mathcal{L}^{o}(\theta_1;\mathcal{T}_{o})+\mathcal{L}^{i}(\theta_2;\mathcal{T}_{i})+\mathcal{L}^{m}(\theta_1,\theta_2;\mathcal{T}_{m})+\mathcal{L}^{b}(\theta_1,\theta_2;\mathcal{T}_{b}),
\end{align}
where  
\begin{align*}
    \mathcal{L}^{o}(\theta_1;\mathcal{T}_{o})=\frac{1}{|\mathcal{T}_{o}|}\sum_{x\in\mathcal{T}_{o}}\left|\mathcal{F}_0(\hat{u}^{o}_{(0)},x)\right|^2,
\end{align*}
\begin{align*}
   \mathcal{L}^{i}(\theta_2;\mathcal{T}_{i})=\frac{1}{|\mathcal{T}_{i}|}\sum_{\xi\in\mathcal{T}_{i}}\left|\tilde{\mathcal{F}}_0(\hat{u}^{i}_{(0)},\xi)\right|^2,
\end{align*}
\begin{align*}
    \mathcal{L}^{m}(\theta_1,\theta_2;\mathcal{T}_{m})=\left|\hat{u}^{o}_{(0)}(x_0)-\hat{u}^{i}_{(0)}(\xi_0)\right|^2,
\end{align*}
\begin{align*}
\mathcal{L}^{b}(\theta_1,\theta_2;\mathcal{T}_{b})=\frac{1}{|\Gamma^o|}\sum_{x\in\Gamma^o}\left|\mathcal{B}^o(\hat{u}^{o}_{(0)},\varepsilon)\right|^2+\frac{1}{|\Gamma^i|}\sum_{\xi\in\Gamma^i}\left|\mathcal{B}^i(\hat{u}^{i}_{(0)},\varepsilon)\right|^2. 
\end{align*}
Here $|\mathcal{T}_{o}|$, $|\mathcal{T}_{i}|$ ,$|\Gamma^o|$, and $|\Gamma^i|$ denote the size of  the corresponding sets, and $\theta=\{\theta_1,\theta_2\}$.
Automatic differentiation \cite{baydin2018automatic} is employed to compute the differential operators in the loss function (\ref{loss-Leading-PVD-Net}), and the optimal parameters are obtained by by minimizing the loss function:
$$\theta=\arg\min_{\theta}\mathcal{L}(\theta;\mathcal{T}).$$
This problem is typically addressed using gradient-based optimization algorithms such as ADAM \cite{kingma2014adam} or L-BFGS \cite{liu1989limited}.

After training the neural networks, in order to obtain a uniformly valid solution, we use a composite solution of the form
\begin{align}\label{Leading-order PVD-Net com}
    \hat{u}(x;\theta)=\hat{u}^{o}_{(0)}(x)+\hat{u}^{i}_{(0)}\left(\frac{x-x_0}{\varepsilon}\right)-\hat{u}^{o,i}_{(0)},\quad x\in\Omega,
\end{align}
where $\hat{u}^{o,i}_{(0)}=\hat{u}^{o}_{(0)}(x_0)$ (or $\hat{u}^{i}_{(0)}({\xi_0})$) is the Prandtl matching term. Although the inner solution $\hat{u}^{i}_{(0)}\left(\frac{x-x_0}{\varepsilon}\right)$ is defined in the inner region $\Omega_i$, it decays rapidly to $u^{o,i}_{(0)}$ outside this region and can be smoothly extended to $\Omega$. Likewise, the outer solution $\hat{u}^{o}_{(0)}(x)$ can be smoothly extrapolated into the inner region.  The matching term $u^{o,i}_{(0)}$ eliminates the overlapping contribution, ensuring the composite solution remains uniformly valid across the domain. An algorithm for implementation of the proposed leading-order PVD-Net is provided in Section SM2.

Furthermore, for the inverse problem of inferring the stretching exponent $\lambda$, an additional loss term is introduced, given by
\begin{align}
    \mathcal{L}^{data}(\theta_2,\lambda;\mathcal{T}_{data})=\frac{1}{|\mathcal{T}_{data}|}\sum_{x\in\mathcal{T}_{data}}\left|\hat{u}^{i}_{(0)}(\frac{x-x_0}{\epsilon^\lambda})-u(x)\right|^2,
\end{align}
where $\mathcal{T}_{data}=\{(x^{(1)},u(x^{(1)})),\dots,(x^{(N)},u(x^{(N)}))\}$ denotes the data available within the boundary layer.
More details are provided in the numerical examples in Section~\ref{inverse}.

\subsubsection{High-order PVD-Net}
The leading-order PVD-Net is suitable for scenarios that prioritize stability. In contrast, Van Dyke’s matching principle accommodates higher-order asymptotic terms, enabling significantly improved solution accuracy when such precision is required. To this end, we propose the high-order PVD-Net, which incorporates Van Dyke’s matching principle to capture higher-order effects more accurately.

Assuming the governing equation is second-order, and in light of resonance phenomena, we employ five fully connected neural networks to construct a high-order surrogate for the solution to Equation~(\ref{Equation}). Specifically, $\hat{u}^o_{(0)}(x;\theta_1)$ and $\hat{u}^i_{(0)}(\xi;\theta_2)$ correspond to the leading-order approximations, while $\hat{u}^o_{(1)}(x;\theta_3)$, $\hat{u}^i_{(c)}(\xi;\theta_4)$, and $\hat{u}^i_{(1)}(\xi;\theta_5)$ are used to model the first-order approximations. Similarly, we simplify notation by using $\hat{u}^o_{(0)}(x)$ and $\hat{u}^i_{(0)}(\xi)$ for the leading-order approximations, and $\hat{u}^o_{(1)}(x)$, $\hat{u}^i_{(c)}(\xi)$, and $\hat{u}^i_{(1)}(\xi)$ for the first-order approximations, whenever the parameter dependence is clear from context.
We propose a novel construction of the two-term asymptotic expansions for both the inner and outer solutions, which can be expressed as:
\begin{align*}
\hat{u}^{o}(x;\theta_1,\theta_3)&=\hat{u}^o_{(0)}(x)+\varepsilon \hat{u}^o_{(1)}(x), \\
\hat{u}^{i}(\xi ;\theta_2,\theta_4,\theta_5)&=\hat{u}^i_{(0)}(\xi)+\varepsilon\{\xi \hat{u}^i_{(c)}(\xi )+\hat{u}^i_{(1)}(\xi)\}.
\end{align*}

\begin{remark}
    The network $\hat{u}^i_{(c)}(\xi)$ is introduced to enforce consistency between the inner and outer approximations under Van Dyke’s matching principle. Since $\xi = \frac{x - x_0}{\varepsilon}$, it follows that $x = \varepsilon \xi + x_0$, and hence the term $\varepsilon \xi \hat{u}^i_{(c)}(\xi)$ can be rewritten as $(x - x_0)\hat{u}^i_{(c)}(\xi)$. This implies that although $\hat{u}^i_{(c)}(\xi)$ appears as part of the first-order  expansion, it contributes to the leading-order behavior. Such design is crucial for realizing Van Dyke’s matching principle, aligning the asymptotic orders between the inner and outer expansions.
\end{remark}

Based on Van Dyke's matching principle and the high-order construction of PVD-Net, we establish the following theorem.

\begin{theorem}\label{th1}
Let $\hat{u}^{o}_{(0)}(\cdot;\theta_1):\Omega_o\times\Theta\to\mathbb{R}$, $\hat{u}^{i}_{(0)}(\cdot;\theta_2):\Omega_i'\times\Theta\to\mathbb{R}$, $\hat{u}^{o}_{(1)}(\cdot;\theta_3):\Omega_o\times\Theta\to\mathbb{R}$, $\hat{u}^{i}_{(c)}(\cdot;\theta_4):\Omega_i'\times\Theta\to\mathbb{R}$, and $\hat{u}^{i}_{(1)}(\cdot;\theta_5):\Omega_i'\times\Theta\to\mathbb{R}$ be five independent neural networks representing the leading-order outer approximation network, leading-order inner approximation network, first-order outer approximation network, order-reduction network, and first-order inner approximation network, respectively. Here, $\Omega_o$ and $\Omega_i'$ denote the outer and stretched inner domains, respectively, $\Theta$ denotes the parameter space, and $\theta_j\in\Theta$, $j=1,\ldots,5$, denotes the trainable parameters of each network. Let $x_0$ be the boundary point at which the boundary layer occurs and $\xi=(x-x_0)/\varepsilon$ be the corresponding stretched variable. Assume that the relevant inner limits exist as $\xi\to+\infty$. Then, the Van Dyke's matching condition for the neural networks is
\[
\resizebox{\linewidth}{!}{$\displaystyle
\lim_{\xi\to+\infty}\hat{u}^{i}_{(0)}(\xi;\theta_2)
=\hat{u}^{o}_{(0)}(x_0;\theta_1),\quad
\lim_{\xi\to+\infty}\hat{u}^{i}_{(c)}(\xi;\theta_4)
=[\hat{u}^o_{(0)}]^{\prime}(x_0;\theta_1),\quad
\lim_{\xi\to+\infty}\hat{u}^{i}_{(1)}(\xi;\theta_5)
=\hat{u}^{o}_{(1)}(x_0;\theta_3),
$}
\]
where $[\hat{u}^o_{(0)}]^{\prime}$ denotes the derivative of
$\hat{u}^o_{(0)}$ with respect to $x$.
\end{theorem}
\noindent The detailed proof is provided in Section~SM1 of the Supplementary Materials.
 
The neural networks' parameters $\theta=\{\theta_1,\theta_2,\theta_3,\theta_4,\theta_5\}$ are optimized by the following loss function:
\begin{equation}\label{loss-high-PVD-Net}
\mathcal{L}(\theta;\mathcal{T})
=
\mathcal{L}^{o}(\theta_1,\theta_3;\mathcal{T}_{o})
+
\mathcal{L}^{i}(\theta_2,\theta_4,\theta_5;\mathcal{T}_{i})
+
\mathcal{L}^{m}(\theta;\mathcal{T}_{m})
+
\mathcal{L}^{b}(\theta;\mathcal{T}_{b})
\end{equation}
where 
\begin{align*}\label{loss-high-PVD-Net-out}
    \mathcal{L}^{o}(\theta_1,\theta_3;\mathcal{T}_{o})=\frac{1}{|\mathcal{T}_{o}|}\sum_{x\in\mathcal{T}_{o}}\left(\left|\mathcal{F}_0(\hat{u}^o_{(0)},x)\right|^2+\left|\mathcal{F}_1(\hat{u}^o_{(0)},\hat{u}^o_{(1)},x)\right|^2\right),
\end{align*}
\begin{align*}\label{loss-high-PVD-Net-in}
      \mathcal{L}^{i}(\theta_2,\theta_4,\theta_5;\mathcal{T}_{i})=\frac{1}{|\mathcal{T}_{i}|}\sum_{x\in\mathcal{T}_{i}}\left|\tilde{\mathcal{F}}(\hat{u}^i,\xi,\varepsilon)\right|^2,
\end{align*}
\begin{equation*}\label{loss-high-PVD-Net-match}
\mathcal{L}^{m}(\theta;\mathcal{T}_{m})
=\left|\hat{u}^o_{(0)}(x_0)-\hat{u}^i_{(0)}(\xi_0)\right|^2
+\left|\hat{u}^o_{(1)}(x_0)-\hat{u}^i_{(1)}(\xi_0)\right|^2
+\left|[\hat{u}^o_{(0)}]^\prime(x_0)-\hat{u}^i_{(c)}(\xi_0)\right|^2
\end{equation*}
\begin{align*}\label{loss-high-PVD-Net-bc}
\begin{aligned}
        \mathcal{L}^{b}(\theta;\mathcal{T}_{b})=\frac{1}{|\Gamma^o|}\sum_{x\in\Gamma^o}\left|\mathcal{B}^o(\hat{u}^{o},\varepsilon)\right|^2+\frac{1}{|\Gamma^i|}\sum_{\xi\in\Gamma^i}\left|\mathcal{B}^i(\hat{u}^{i},\varepsilon)\right|^2.
\end{aligned}
\end{align*}

\begin{remark}
    From the perspective of well-posedness, the inner and outer approximations require different treatments in the loss design. The outer expansions $\hat{u}^o_{(0)}$ and $\hat{u}^o_{(1)}$ correspond to separate orders of the original governing equation posed on the outer domain, where each order often admits a well-defined boundary value problem. Thus, it is both natural and mathematically justified to impose separate residual losses $\mathcal{F}_0$ and $\mathcal{F}_1$ for the leading and first-order outer approximations.  In contrast,  the inner asymptotic expansion terms $\hat{u}^i_{(0)}$, $\hat{u}^i_{(1)}$ and $\hat{u}^i_{(c)}$ do not form well-posed problems on their own. To ensure well-posedness and stable training, we treat the inner approximation as a unified whole and define the residual loss based on the transformed  equation $\tilde{\mathcal{F}}(u^i,\xi,\varepsilon)$. This approach constrains all inner components simultaneously while preserving the structure of the asymptotic expansion.
\end{remark}

Finally, we construct the composite solution in the form: 
\begin{align}\label{com-high-order PVD-Net}
  \hat{u}(x;\theta)=\hat{u}^{o}(x)+\hat{u}^{i}\left(\frac{x-x_0}{\varepsilon}\right)-\hat{u}^{o,i}_{(1)},\quad x\in\Omega,
\end{align}
where $\hat{u}^{o,i}_{(1)}=\hat{u}^i_{(0)}(\xi_0)+\hat{u}^i_{(c)}(\xi_0)(x-x_0)+\varepsilon\hat{u}^i_{(1)}(\xi_0)$
is the Van Dyke matching term. Following the same strategy as described in leading-order PVD-Net, the inner and outer solutions are extended across $\Omega$, and the overlapping contribution is removed by the matching term. The algorithmic implementation of the proposed high-order PVD-Net is outlined in Section SM2.

\subsection{Prandtl-Van Dyke Deep Operator Network (PVD-ONet)}
In this section, we introduce PVD-ONet, which extends PVD-Net through operator learning to approximate solution operators for families of boundary layer problems. Both leading-order and high-order formulations are developed, providing a balance between stability and accuracy.
\subsubsection{Physics-informed DeepONet}
The DeepONet \cite{lu2021learning} is a powerful framework for learning solution operators of entire families of PDEs. Let 
$\mathcal{V}$ denotes the input function space, typically containing initial conditions, boundary conditions, or other parameters, and let $\mathcal{U}$ denotes the solution space of the corresponding PDEs. We consider a general family of parametric partial differential equations of the form
\begin{align*}
    \mathcal{N}(v,u)=0,
\end{align*}
where $v\in\mathcal{V}$ denotes the input function, and $u\in\mathcal{U}$ denotes the corresponding solution of the system. The goal is to learn the solution operator $G_\theta:\mathcal{V}\to \mathcal{U}$, which maps each input $v\in \mathcal{V}$
to its associated solution $u=G_\theta(v)\in\mathcal{U}$.
To represent $v$ numerically, it is necessary to discretize it using a set of ``sensor points'' $\{x_1,x_2,\dots,x_M\}\subset \Omega$. The function $v$ is then approximated by its point-wise evaluations at these sensor points:
\begin{align*}
    \boldsymbol{v}=[v(x_1),v(x_2),\dots,v(x_M)]\in\mathbb{R}^M,
\end{align*}
where $\boldsymbol{v}$ is the finite-dimensional representation of $v$. The DeepONet consists of two main sub-networks: the ``branch" network and the ``trunk" network. The discretized input $v$ is fed into the branch network, which encodes it into a finite-dimensional feature vector. The trunk network takes the coordinate 
$\zeta\in \Omega$ as input. Finally, the output of the DeepONet is denoted as
\begin{align*}
    G_\theta(v)(\zeta):=\sum_{i=1}^p b_i(\boldsymbol{v})\cdot t_i(\zeta),
\end{align*}
where $b_1,b_2,\dots,b_p$ is the output of the branch network, $t_1,t_2,\dots,t_p$ is the output of the trunk network. 

\begin{figure}[h]
    \centering
    \includegraphics[width=0.6\linewidth]{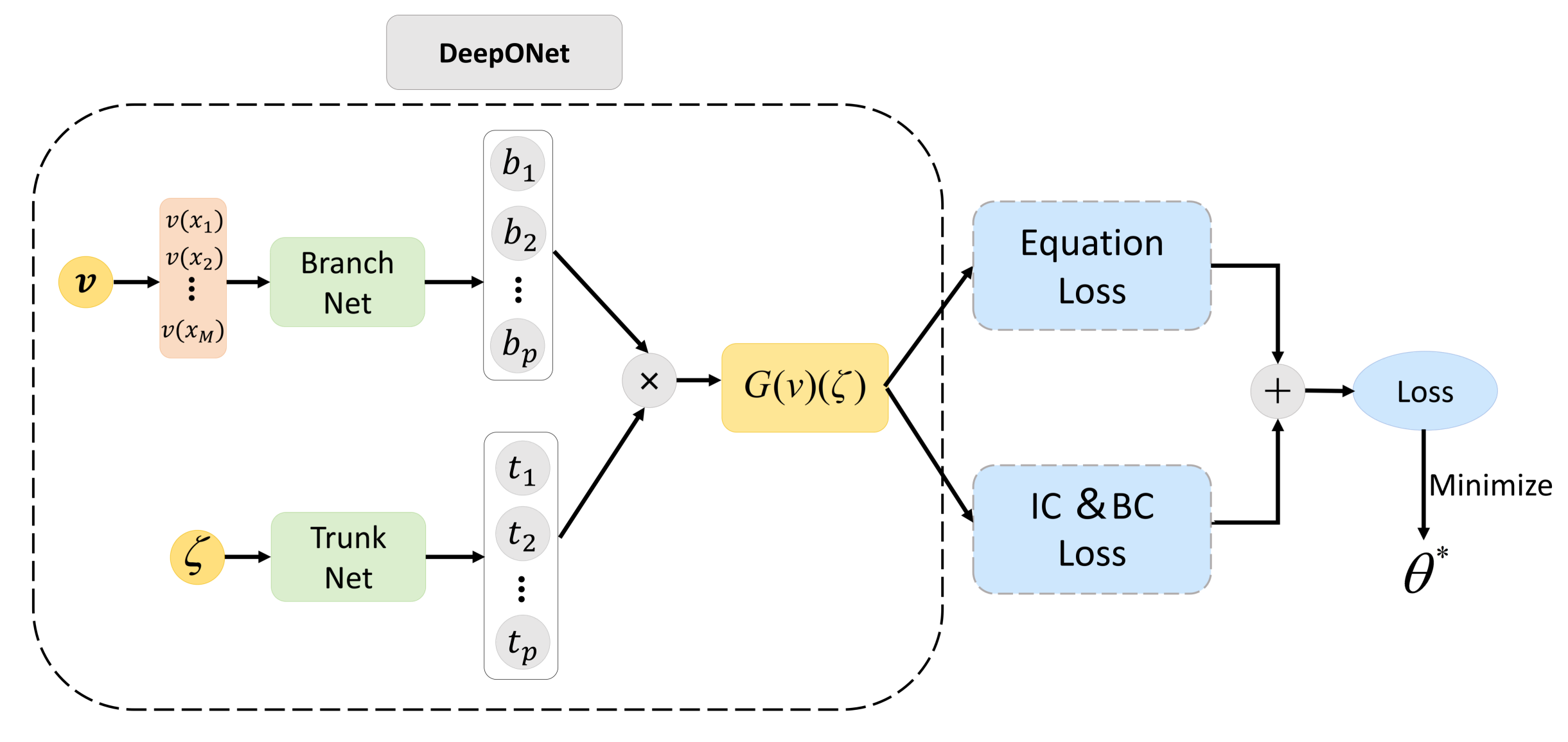}
    \caption{Architecture of the physics-informed DeepONet. The branch network encodes the input function $v$ sampled at fixed sensor locations ${x_1,x_2,\ldots,x_M}$, while the trunk network encodes the coordinate $\zeta$. Their outputs are combined through a dot product to approximate the solution operator. Automatic differentiation evaluates the PDE residuals and enforces the physical constraints.}
    \label{deeponet}
\end{figure}

Building upon the original DeepONet framework, Physics-informed DeepONet \cite{wang2021learning} further integrates physical knowledge in the loss function. Specifically, the loss function is defined as follows:
\begin{align*}
\mathcal{L}=\frac{1}{NJM}\sum_{n=1}^N\sum_{j=1}^J\sum_{m=1}^M\left|\mathcal{N}(v_n(x_m),G_\theta(v_n)(\zeta_{n,j}))\right|^2,
\end{align*} 
where $\{v_n\}_{n=1}^N\subset\mathcal{V}$ are input functions sampled from the input space, for each $n$, $\{\zeta_{n,j}\}_{j=1}^J$ denotes a set of collocation points sampled within the domain where $G_\theta(v_n)$ is defined. $\{x_m\}_{m=1}^M$ represents a fixed set of sensor points used to evaluate the input functions.
The overall architecture of the Physics-informed DeepONet is depicted in Figure \ref{deeponet}.

\subsubsection{Leading-order PVD-ONet}
In this subsection, we extend the leading-order PVD-Net framework to the operator learning setting by incorporating the DeepONet architecture.  The inner and outer networks, denoted as $G_{(0)}^{o}:\mathcal{V}\to\mathcal{U}$ and $G_{(0)}^{i}:\mathcal{V}\to\mathcal{U}$, are parameterized using DeepONets with parameters $\theta_1$ and $\theta_2$,
respectively. They are designed to approximate the inner solution and outer solution of the parametric boundary layer problem.
The Prandtl matching principle remains essentially the same as leading-order PVD-Net, is expressed as
\begin{align*}
    \lim_{x\to x_0}G^{o}_{(0)}(v)(x)=\lim_{\xi \to\infty}G^{i}_{(0)}(v)(\xi),\quad v\in\mathcal{V}.
\end{align*}

The loss function for the leading-order PVD-ONet is structured similarly to the leading-order PVD-Net. Specifically, for $\theta=\{\theta_1,\theta_2\}$, the loss function is defined as follows:
\begin{align}\label{loss-Leading-PVD-ONet}
    \mathcal{L}(\theta;\mathcal{T}):=\mathcal{L}^{o}(\theta_1;\mathcal{T}_{o})+\mathcal{L}^{i}(\theta_2;\mathcal{T}_{i})+\mathcal{L}^{m}(\theta_1,\theta_2;\mathcal{T}_{m})+\mathcal{L}^{b}(\theta_1,\theta_2;\mathcal{T}_{b}),
\end{align}
where  
\begin{align*}\label{loss-Leading-PVD-ONet-out}
    \mathcal{L}^{o}(\theta_1;\mathcal{T}_{o})=\frac{1}{N|\mathcal{T}_{o}|}\sum_{n=1}^N\sum_{x\in\mathcal{T}_{o}}\left|\mathcal{F}_0(G^{o}_{(0)}(v_n),x)\right|^2,
\end{align*}
\begin{align*}\label{loss-Leading-PVD-oNet-in}
   \mathcal{L}^{i}(\theta_2;\mathcal{T}_{i})=\frac{1}{N|\mathcal{T}_{i}|}\sum_{n=1}^N\sum_{\xi\in\mathcal{T}_{i}}\left|\tilde{\mathcal{F}}_0(G^{i}_{(0)}(v_n),\xi)\right|^2,
\end{align*}
\begin{align*}\label{loss-Leading-PVD-ONet-match}
    \mathcal{L}^{m}(\theta_1,\theta_2;\mathcal{T}_{m})=\frac{1}{N}\sum_{n=1}^N\left|G^{o}_{(0)}(v_n)(x_o)-G^{i}_{(0)}(v_n)(\xi_0)\right|^2,
\end{align*}
\begin{align*}\label{loss-Leading-PVD-ONet-bc}
\mathcal{L}^{b}(\theta_1,\theta_2;\mathcal{T}_{b})=\frac{1}{N|\Gamma^o|}\sum_{n=1}^N\sum_{x\in\Gamma^o}\left|\mathcal{B}^o(G^{o}_{(0)}(v_n),\varepsilon)\right|^2+\frac{1}{N|\Gamma^i|}\sum_{n=1}^N\sum_{\xi\in\Gamma^i}\left|\mathcal{B}^i(G^{i}_{(0)}(v_n),\varepsilon)\right|^2.
\end{align*}

The composite solution for the uniformly valid approximation is given by:
\begin{align}\label{Leading-order-PVD-ONet-com}
    G(v_n)(x;\theta) = G^{o}_{(0)}(v_n)(x) + G^{i}_{(0)}(v_n)\left(\frac{x-x_0}{\varepsilon}\right) - G^{o,i}_{(0)}(v_n),\quad x\in\Omega,
\end{align}
where $G^{o,i}_{(0)}(v_n) =G_{(0)}^o(v_n)(x_0)$ (or $G_{(0)}^i(v_n)(\xi_0)$) is the Prandtl matching term.


\subsubsection{High-order PVD-ONet}
We extend the high-order PVD-Net framework by incorporating the DeepONet architecture to enable operator learning. As a result, both the outer and inner solutions are now approximated through DeepONet. Specifically, analogous to the definitions of the high-order PVD-Net,  $G^o_{(0)}:\mathcal{V}\to\mathcal{U}$  and $G^i_{(0)}:\mathcal{V}\to\mathcal{U}$, parameterized by $\theta_1$ and $\theta_2$, respectively, represent the leading-order operator approximations. The first-order operator approximations are modeled by 
 $G^o_{(1)}:\mathcal{V}\to\mathcal{U}$, $G^i_{(c)}:\mathcal{V}\to\mathcal{U}$ and $G^i_{(1)}:\mathcal{V}\to\mathcal{U}$ with corresponding parameters $\theta_3$, $\theta_4$
 and $\theta_5$.
 We propose a novel operator-learning-based framework for constructing two-term asymptotic expansions of both the inner and outer solutions, which can be expressed as:
\begin{align*}
G^{o}(v)(x;\theta_1,\theta_3)&=G^o_{(0)}(v)(x)+\varepsilon G^o_{(1)}(v)(x), \\
G^{i}(v)(\xi;\theta_2,\theta_4,\theta_5)&=G^i_{(0)}(v)(\xi)+\varepsilon\{\xi G^i_{(c)}(v)(\xi)+G^i_{(1)}(v)(\xi)\}.
\end{align*}
Similar to the high-order PVD-Net's setting, the neural networks' parameters $\theta=\{\theta_1,\theta_2,\theta_3,\theta_4,\theta_5\}$ are optimized by the following loss function:
\begin{align}\label{loss-high-PVD-ONet}
    \mathcal{L}(\theta;\mathcal{T})=\mathcal{L}^{o}(\theta_1,\theta_3;\mathcal{T}_{o})+
    \mathcal{L}^{i}(\theta_2,\theta_4,\theta_5;\mathcal{T}_{i})+\mathcal{L}^{m}(\theta;\mathcal{T}_{m})+\mathcal{L}^{b}(\theta;\mathcal{T}_{b}),
\end{align}
where 
\begin{align*}
    \mathcal{L}^{o}(\theta_1,\theta_3;\mathcal{T}_{o})=\frac{1}{N|\mathcal{T}_{o}|}\sum_{n=1}^N\sum_{x\in\mathcal{T}_{o}}\left(\left|\mathcal{F}_0(G^o_{(0)}(v_n),x)\right|^2+\left|\mathcal{F}_1(G^o_{(0)}(v_n),G^o_{(1)}(v_n),x)\right|^2\right),
\end{align*}
\begin{align*}
      \mathcal{L}^{i}(\theta_2,\theta_4,\theta_5;\mathcal{T}_{i})=\frac{1}{N|\mathcal{T}_{i}|}\sum_{n=1}^N\sum_{x\in\mathcal{T}_{i}}\left|\tilde{\mathcal{F}}(G^i(v_n),\xi,\varepsilon)\right|^2,
\end{align*}
\begin{align*}
\mathcal{L}^{m}(\theta;\mathcal{T}_{m})
&=\frac{1}{N}\sum_{n=1}^{N}\Big(
\left|G^o_{(0)}(v_n)(x_0)-G^i_{(0)}(v_n)(\xi_0)\right|^2
+\left|G^o_{(1)}(v_n)(x_0)-G^i_{(1)}(v_n)(\xi_0)\right|^2 \\
&\qquad\qquad
+\left|[G^o_{(0)}(v_n)]^\prime(x_0)-G^i_{(c)}(v_n)(\xi_0)\right|^2
\Big).
\end{align*}
\begin{align*}
\mathcal{L}^{b}(\theta;\mathcal{T}_{b})=\frac{1}{N|\Gamma^o|}\sum_{n=1}^N\sum_{x\in\Gamma^o}\left|\mathcal{B}^o(G^{o}(v_n),\varepsilon)\right|^2+\frac{1}{N|\Gamma^i|}\sum_{n=1}^N\sum_{\xi\in\Gamma^i}\left|\mathcal{B}^i(G^{i}(v_n),\varepsilon)\right|^2.
\end{align*}
Finally, the composite solution is constructed as:
\begin{align}
  G(v_n)(x;\theta)=G^o(v_n)(x)+G^i(v_n)\left(\frac{x-x_o}{\varepsilon}\right)-G^{o,i}_{(1)}(v_n),\quad x\in \Omega, 
\end{align}
where $G^{o,i}_{(1)}(v_n)=G_{(0)}^i(v_n)(\xi_0)+G_{(c)}^i(v_n)(\xi_0)(x-x_0)+\varepsilon G_{(1)}^i(v_n)(\xi_0)$.

\section{Numerical experiments}\label{numerical experiments}
In this section, we evaluate the proposed PVD-Net and PVD-ONet through numerical experiments. Unless otherwise specified, the hyperparameter configurations used in all numerical experiments are summarized in Table~\ref{hyperparameter}. Additional details on network depth, width, number of networks, trainable parameters, and training epochs are provided in the supplementary material SM3. Experiments are performed on an NVIDIA RTX A6000 48GB GPU.

\begin{table}
\centering
\caption{Training hyperparameter settings for all methods.}
\label{hyperparameter}
\begin{tabular}{cc}
\toprule
\textbf{Item} & \textbf{Setting} \\
\midrule
Optimizer & Adam \\
Learning rate & $10^{-4}$ \\
Training epochs & 100{,}000 \\
Initialization & Kaiming (He) \\
Activation & SiLU \\
Model selection & Best checkpoint \\
Truncation parameter $\xi_0$ & 20 \\
\bottomrule
\end{tabular}
\end{table}

\subsection{Accuracy Evaluation Metrics}

To quantitatively evaluate the accuracy of the predicted solutions, we consider
the relative $L^2$ error and the $L^\infty$ error, defined as
\begin{align*}
\mathcal{E}_{L^2}
&=
\frac{
\left(\sum_{i=1}^{N}
\left|\hat{u}(x_i,y_i)-u(x_i,y_i)\right|^2
\right)^{1/2}
}{
\left(\sum_{i=1}^{N}
\left|u(x_i,y_i)\right|^2
\right)^{1/2}
},
\qquad
\mathcal{E}_{L^\infty}
=
\max_{1\leq i\leq N}
\left|\hat{u}(x_i,y_i)-u(x_i,y_i)\right|.
\end{align*}
Here, $\hat{u}$ and $u$ denote the predicted solution and the reference solution,
respectively, and $\{(x_i,y_i)\}_{i=1}^{N}$ represents the set of evaluation
points.

\subsection{Second-order equation with constant coefficients}
Let us consider a second-order equation with constant coefficients presented in
\begin{align}\label{case1}
 \left\{
\begin{aligned}
    \varepsilon \frac{d^{2} u}{dx^{2}}&+\frac{d u}{d x}+u=0,\quad x\in(0,1),\\
    u(0)&=\alpha,\quad u(1)=\beta,
\end{aligned}
\right.
\end{align}
where $\varepsilon=10^{-3}$, $\alpha=1$ and $\beta=2$. Equation (\ref{case1}) has an analytical solution given by 
\begin{align}\label{2}
 u=\frac{-\alpha e^{\lambda_{2}}+\beta}{e^{\lambda_{1}}-e^{\lambda_{2}}}e^{\lambda_{1}x}+\frac{\alpha e^{\lambda_{1}}-\beta}{e^{\lambda_{1}}-e^{\lambda_{2}}}e^{\lambda_{2}x},
\end{align}
where $\lambda_{1}=\frac{-1+\sqrt{1-4\varepsilon}}{2\varepsilon}$, $\lambda_{2}=\frac{-1-\sqrt{1-4\varepsilon}}{2\varepsilon}$. This equation represents a fundamental problem in the theory of singular perturbations \cite{nayfeh2024perturbation}. 
In this example, the boundary layer is at $x=0$, and an inner expansion using the stretching transformation $\xi =\frac{x}{\varepsilon}$ was introduced to amplify the boundary layer. A detailed analysis of this example is provided in the Section SM4, where the coefficients are set as $a(x)=1$, $b(x)=1$. We also consider a second-order problem with variable coefficients to further evaluate the proposed method. Due to space limitations, the corresponding results are provided in the Section SM5.
\paragraph{PVD-Net}
For this boundary layer problem, the leading-order PVD-Net uses two networks to approximate the outer and inner solutions, whereas the high-order PVD-Net employs five networks, including three additional networks for the first-order correction. The corresponding loss formulations are given in Section~SM6.1. For evaluation, 10,000 points are sampled in the inner region, 100 in the outer region, and one at the interface. Relative $L^2$ and $L^\infty$ errors are computed in the inner region, while the global errors are evaluated over all 10,101 points.

\begin{figure}[ht]
    \centering
    \begin{subfigure}[t]{0.48\linewidth}
        \centering
        \includegraphics[width=\linewidth]{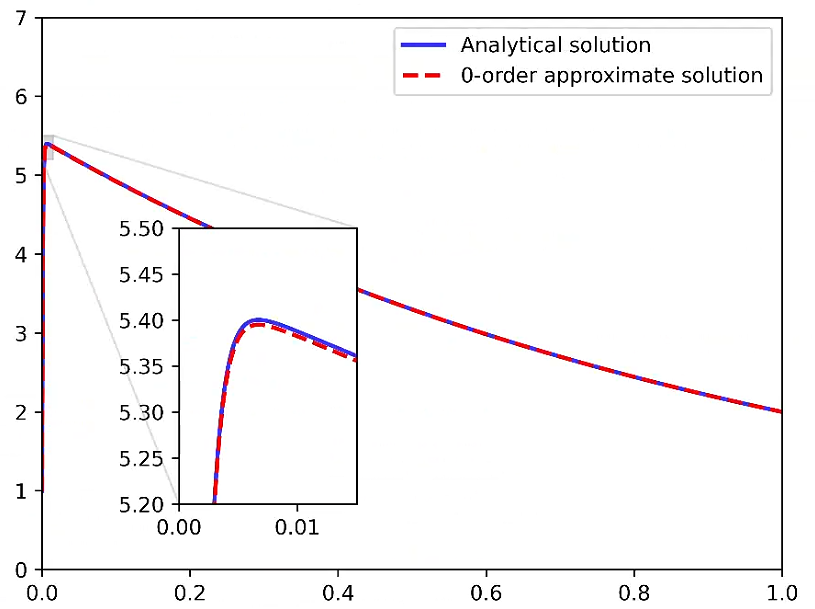}
    \end{subfigure}
    \hfill
    \begin{subfigure}[t]{0.48\linewidth}
        \centering
        \includegraphics[width=\linewidth]{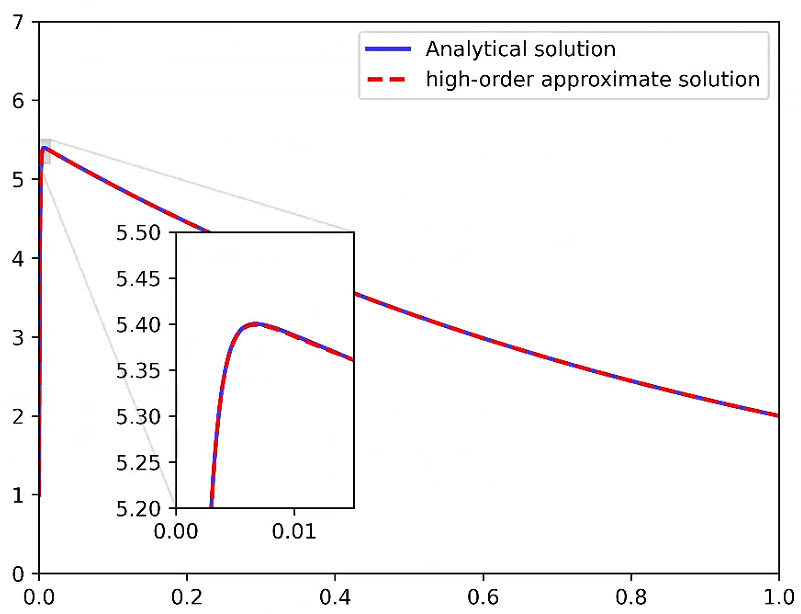}
    \end{subfigure}
    \caption{\textbf{(Left):} Prediction of the leading-order PVD-Net. \textbf{(Right):} Prediction of the high-order PVD-Net. Both versions accurately capture the boundary layer solution with smooth transitions. The high-order PVD-Net further improves accuracy over the leading-order version, particularly in the boundary layer and transition regions.}
    \label{case1-PVD-Net}
\end{figure}

Figure~\ref{case1-PVD-Net} shows that both PVD-Net accurately resolve the boundary layer and maintain smooth predictions across the domain. In particular, the high-order PVD-Net yields substantially higher accuracy, with a pronounced improvement in the boundary layer transition region. As reported in Table~\ref{table-pvd-net-const}, the high-order PVD-Net achieves the lowest errors across all metrics, clearly outperforming the leading-order formulation and BL-PINNs. Compared with MSM-NN, it attains comparable relative $L^2$ errors while achieving smaller $L^\infty$ errors, further demonstrating its superior ability to resolve sharp boundary layer structures.

\begin{table}[ht]
\centering
\caption{Comparison of prediction errors and training time for PVD-Net on the constant-coefficient problem. The bolded values indicate the lowest errors among all methods.}
\label{table-pvd-net-const}
\resizebox{\textwidth}{!}{
\begin{tabular}{cccccccc}
\toprule
\textbf{Method}
& \multicolumn{2}{c}{\textbf{Global Errors}}
& \multicolumn{2}{c}{\textbf{Inner Region Errors}}
& \textbf{Junction Point}
& \textbf{Training} \\
\cmidrule(r){2-3} \cmidrule(r){4-5}
& Relative $L^2$
& $L^\infty$
& Relative $L^2$
& $L^\infty$
& \textbf{Error}
& \textbf{Time (s)} \\
\midrule

BL-PINNs \cite{arzani2023theory}
    & $1.50\times10^{-3}$ & $2.27\times10^{-2}$
    & $1.50\times10^{-3}$ & $2.27\times10^{-2}$
    & $2.21\times10^{-2}$
    & 655 \\

MSM-NN \cite{zhang2024multi}
    & $5.54\times10^{-4}$ & $8.94\times10^{-3}$
    & $5.55\times10^{-4}$ & $8.94\times10^{-3}$
    & $3.03\times10^{-4}$
    & 680 \\

Leading-order PVD-Net
    & $9.43\times10^{-4}$ & $5.39\times10^{-3}$
    & $9.44\times10^{-4}$ & $5.39\times10^{-3}$
    & $5.20\times10^{-3}$
    & 766 \\

\textbf{High-order PVD-Net}
    & $\mathbf{1.44\times10^{-4}}$ & $\mathbf{1.90\times10^{-3}}$
    & $\mathbf{1.44\times10^{-4}}$ & $\mathbf{1.90\times10^{-3}}$
    & $\mathbf{1.02\times10^{-4}}$
    & 1,380 \\

\bottomrule
\end{tabular}
}
\end{table}

\begin{figure}[h]
    \centering
    \includegraphics[width=0.80\linewidth]{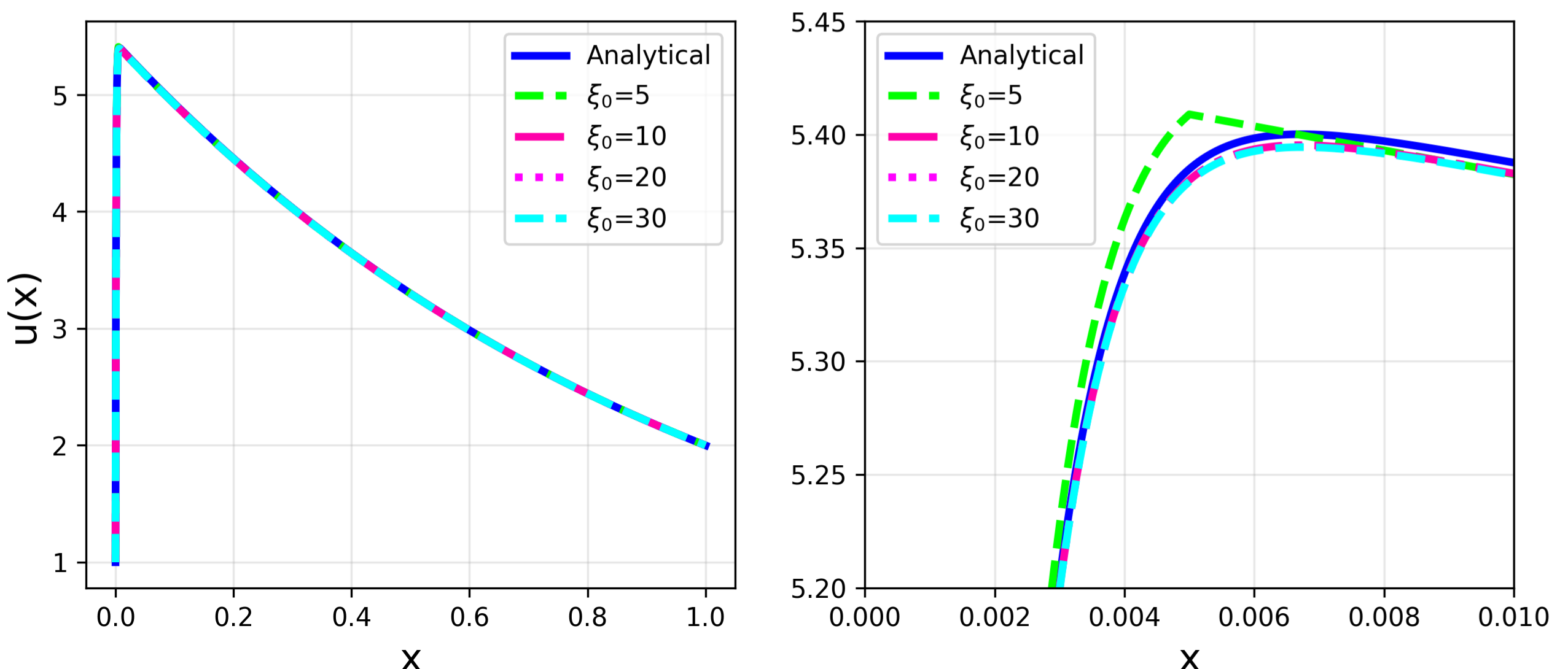}
    \caption{\textbf{Sensitivity to the truncation parameter $\xi_0$ for leading-order PVD-Net.} (Left): reconstructed solutions for $\xi_0 = 5, 10, 20, 30$. 
(Right): zoomed-in views near the boundary layer. A too small truncation ($\xi_0 = 5$) leads to visible errors, while larger values yield nearly identical and accurate predictions.}
    \label{fig:A}
\end{figure}

To assess the sensitivity of the method to the truncation parameter $\xi_0$, we conduct a simple study by varying $\xi_0 = 5, 10, 20, 30$. The results for the leading-order PVD-Net are shown in Figure \ref{fig:A}.  When $\xi_0 = 5$, the prediction exhibits noticeable inaccuracies, indicating that the truncated domain is insufficient to fully capture the layer structure. In contrast, for $\xi_0 = 10, 20, 30$, the predictions are all in close agreement with the reference solution, with negligible differences among them. This suggests that the method is robust with respect to $\xi_0$ once it is chosen sufficiently large. In practice, $\xi_0$ can be selected to ensure that the inner domain fully covers the layer region, beyond which further increases have little impact on the solution accuracy. In this work, we set $\xi_0 = 20$, which is sufficiently large to ensure accurate resolution of the boundary layer.

\begin{figure}[h]
    \centering
    \begin{subfigure}[t]{0.44\linewidth}
        \centering
        \includegraphics[width=\linewidth]{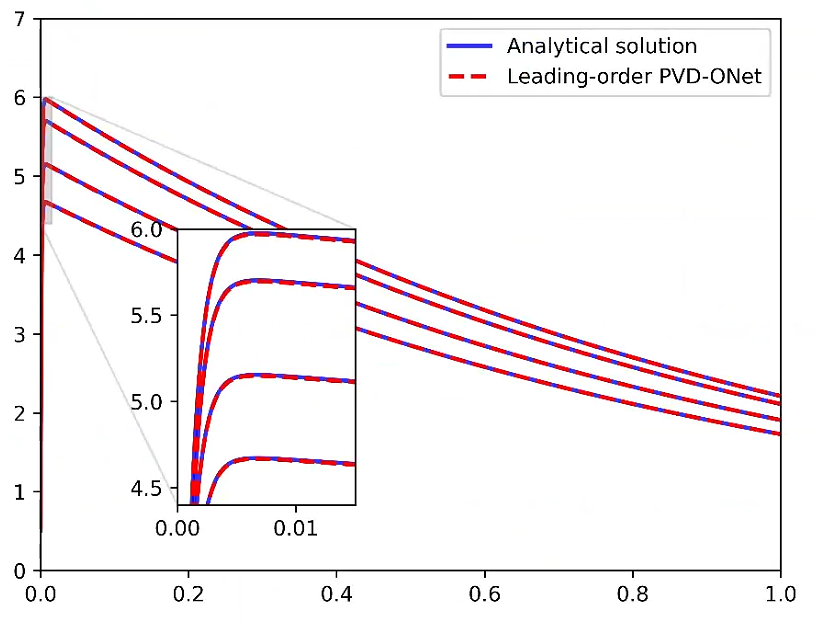}
    \end{subfigure}
    \hfill
    \begin{subfigure}[t]{0.44\linewidth}
        \centering
        \includegraphics[width=\linewidth]{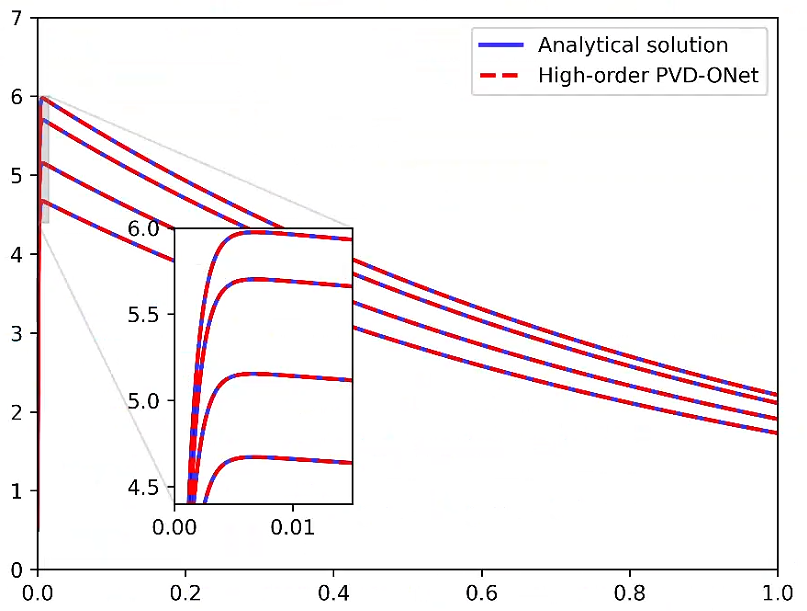}
    \end{subfigure}
    \caption{\textbf{(Left):} Predicted results of Leading-order PVD-ONet. \textbf{(Right):} Predicted results of High-order PVD-ONet. We aim to learn the operator $G:(a,b)\mapsto u$.}
    \label{case1-PVD-ONet}
\end{figure}

\paragraph{PVD-ONet} For PVD-ONet, we learn a solution operator $G:(a,b)\mapsto u$ that maps boundary conditions to solutions, where $a\sim \mathcal{U}(0.4,1.4)$ and $b\sim \mathcal{U}(1.5,2.5)$. We use 1000 boundary conditions for training and 100 for testing. PVD-ONet relies solely on the governing equations without requiring additional observational data. The prediction results are shown in Figure \ref{case1-PVD-ONet}, where both leading-order and high-order PVD-ONet accurately capture the solution with smooth transitions. The high-order PVD-ONet further improves accuracy, particularly near the boundary layer interface. Table \ref{table:pvd-onet-const} compares different methods, demonstrating that PVD-ONet outperforms PI-DeepONet, with the high-order version achieving higher accuracy.

\begin{table}[H]
\centering
\caption{Comparison of mean prediction errors for PVD-ONet on the constant-coefficient problem. Bold indicates the lowest error among all methods.}
\label{table:pvd-onet-const}
\resizebox{\linewidth}{!}{%
\begin{tabular}{cccccc}
\toprule
\textbf{Method} 
& \multicolumn{2}{c}{\textbf{Global Errors}} 
& \multicolumn{2}{c}{\textbf{Inner Region Errors}} 
& \textbf{Training} \\
\cmidrule(r){2-3} \cmidrule(r){4-5}
& Relative $L^2$ & $L^\infty$ 
& Relative $L^2$ & $L^\infty$ 
& \textbf{Time (s)} \\
\midrule
PI-DeepONet \cite{wang2021learning}    
& $8.71 \times 10^{-1}$ & $4.80 \times 10^{0}$ 
& $8.71 \times 10^{-1}$ & $4.80 \times 10^{0}$ & 6,817 \\
Leading-order PVD-ONet                 
& $1.01 \times 10^{-3}$ & $5.83 \times 10^{-3}$ 
& $1.01 \times 10^{-3}$ & $5.83 \times 10^{-3}$ & 18,495 \\
\textbf{High-order PVD-ONet}            
& $\mathbf{1.5 \times 10^{-4}}$ & $\mathbf{1.82 \times 10^{-3}}$ 
& $\mathbf{1.5 \times 10^{-4}}$ & $\mathbf{1.82 \times 10^{-3}}$ & 33,656 \\
\bottomrule
\end{tabular}%
}
\end{table}

\subsection{Internal layer problem}
We consider the internal layer problem as follows:
\begin{align}\label{case3}
 \left\{
\begin{aligned}
    \varepsilon \frac{d^{2} u}{dx^{2}}&=u\frac{d u}{d x}-u,\quad x\in(0,1),\\
    u(0)&=\alpha,\quad u(1)=\beta,
\end{aligned}
\right.
\end{align}
where $\varepsilon=10^{-3}$, $\alpha=1$, $\beta=-1$. The boundary layer is located around $x_0=\frac{1}{2}$, indicating an internal layer phenomenon. 
The reference solution is obtained numerically using a boundary value problem (BVP) solver. 

\paragraph{PVD-Net}
For the leading-order PVD-Net, three networks are employed: one inner network for the internal layer and two outer networks for the left and right regions. The high-order PVD-Net extends this architecture to seven networks, including three inner networks for the leading- and first-order approximations and four outer networks for the corresponding left and right approximations. The detailed formulation and loss construction are provided in Section~SM6.2. For evaluation, 10,000 points are sampled in the internal layer region and 100 points in the outer regions, and both inner-region and global relative $L^2$ and $L^\infty$ errors are reported.

\begin{figure}[h]
    \centering
    \begin{subfigure}[t]{0.46\linewidth}
        \centering
        \includegraphics[width=\linewidth]{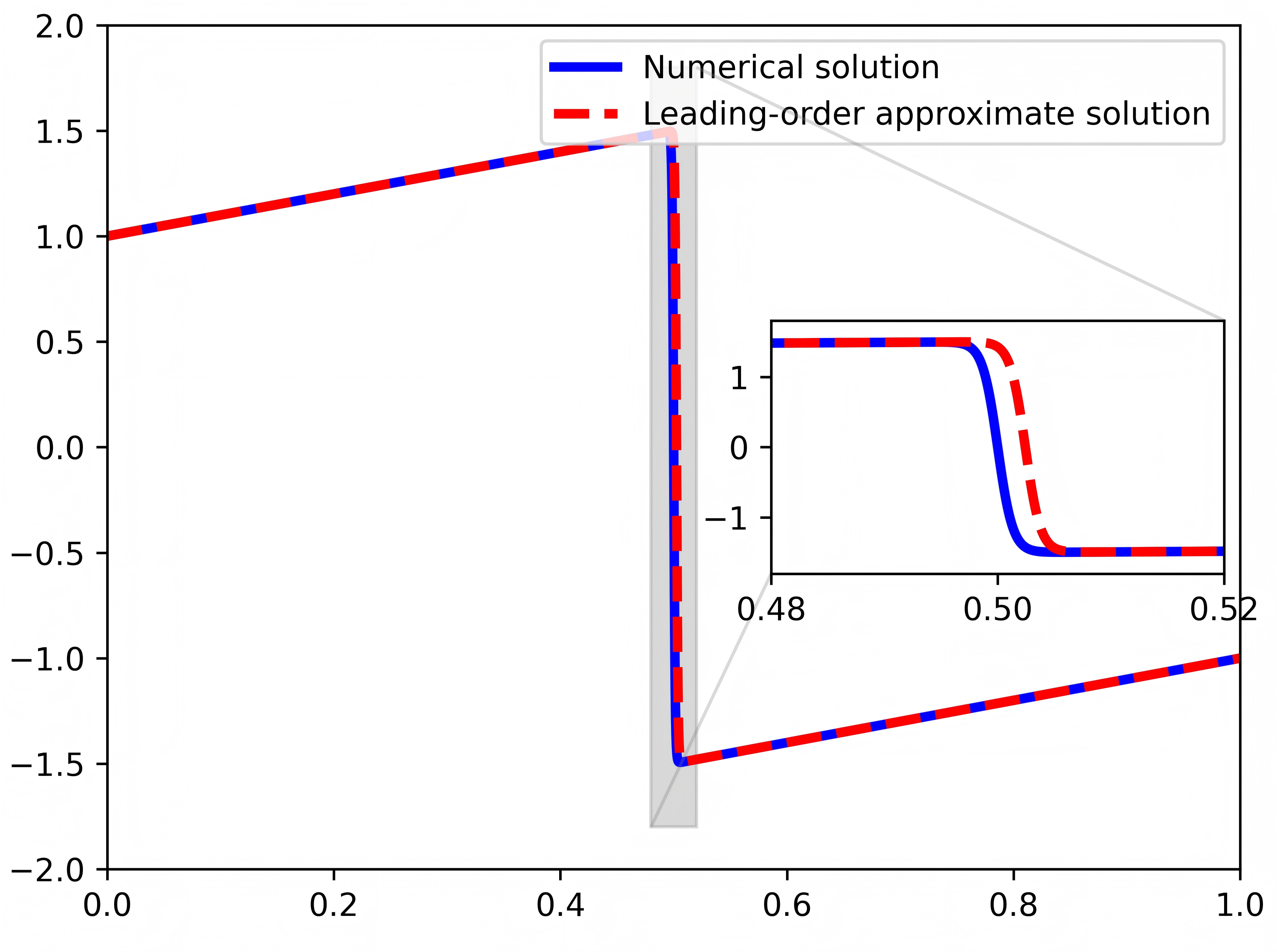}
    \end{subfigure}
    \hfill
    \begin{subfigure}[t]{0.46\linewidth}
        \centering
        \includegraphics[width=\linewidth]{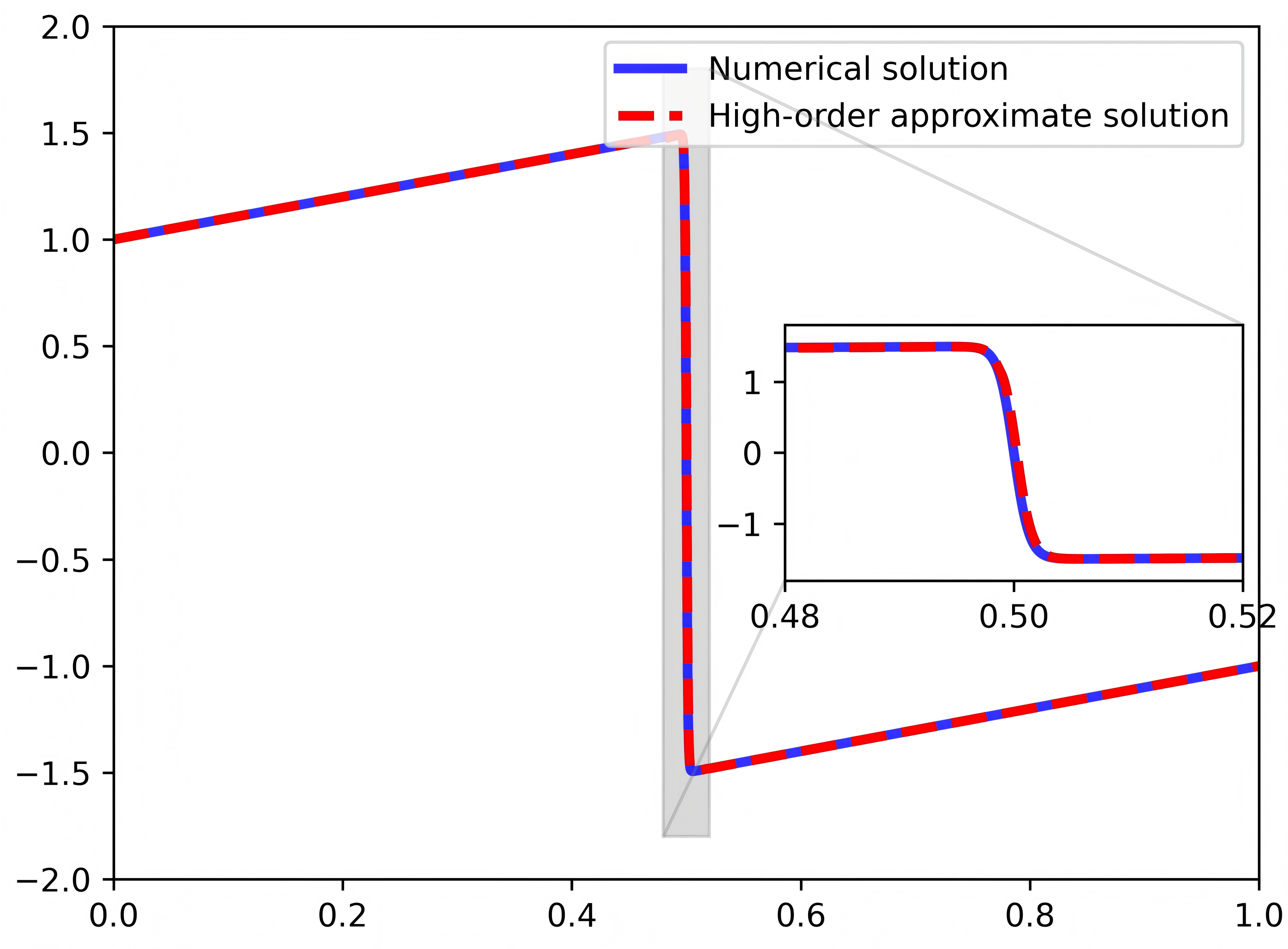}
    \end{subfigure}
    \caption{Prediction results of the PVD-Net for the internal layer problem. \textbf{(Left):} Predicted results of Leading-order PVD-Net. \textbf{(Right):} Predicted results of High-order PVD-Net.  A closer examination through localized zoom-in plots reveals that the high-order PVD-Net achieves superior accuracy within the internal layer region.}
    \label{case3-PVD-Net}
\end{figure}

As shown in Fig.~\ref{case3-PVD-Net}, the leading-order approximation captures the overall trend but shows noticeable deviations in the internal layer region. In contrast, the high-order PVD-Net significantly improves the approximation accuracy and better matches the numerical solution. Table~\ref{table-pvd-net-internal} demonstrates that the high-order PVD-Net outperforms all baseline methods in both global and inner-region errors, reducing the relative $L^2$ error by nearly one order of magnitude compared with the leading-order variant. Although the high-order model requires additional training cost due to multiple subnetworks and loss terms, this overhead enables more accurate resolution of higher-order asymptotic structures, particularly in the sharp internal layer region.

\begin{table}[h]
\centering
\caption{Comparison of prediction errors for PVD-Net on the internal layer problem after training for 100,000 epochs. Bold indicates the lowest error.}
\label{table-pvd-net-internal}
\resizebox{\linewidth}{!}{%
\begin{tabular}{lccccc}
\toprule
\textbf{Method} 
& \multicolumn{2}{c}{\textbf{Global Errors}} 
& \multicolumn{2}{c}{\textbf{Inner Errors}} 
& \textbf{Training} \\
\cmidrule(r){2-3} \cmidrule(r){4-5}
& $L^2$ & $L^\infty$ & $L^2$ & $L^\infty$ 
& \textbf{Time (s)} \\
\midrule
BL-PINNs \cite{arzani2023theory}         
& $1.25\times10^{0}$ & $3.00\times10^{0}$ 
& $1.24\times10^{0}$ & $3.00\times10^{0}$ 
& 644 \\
MSM-NN \cite{zhang2024multi}             
& $9.42\times10^{-1}$ & $1.97\times10^{0}$ 
& $9.48\times10^{-1}$ & $1.97\times10^{0}$ 
& 1,091 \\
Leading-order PVD-Net                   
& $3.64\times10^{-1}$ & $2.17\times10^{0}$ 
& $3.66\times10^{-1}$ & $2.17\times10^{0}$ 
& 984 \\
\textbf{High-order PVD-Net}              
& $\mathbf{4.03\times10^{-2}}$ & $\mathbf{2.82\times10^{-1}}$ 
& $\mathbf{4.06\times10^{-2}}$ & $\mathbf{2.82\times10^{-1}}$ 
& 1,995 \\
\bottomrule
\end{tabular}%
}
\end{table}

\begin{figure}[h]
    \centering
    \begin{subfigure}[t]{0.46\linewidth}
        \centering
        \includegraphics[width=\linewidth]{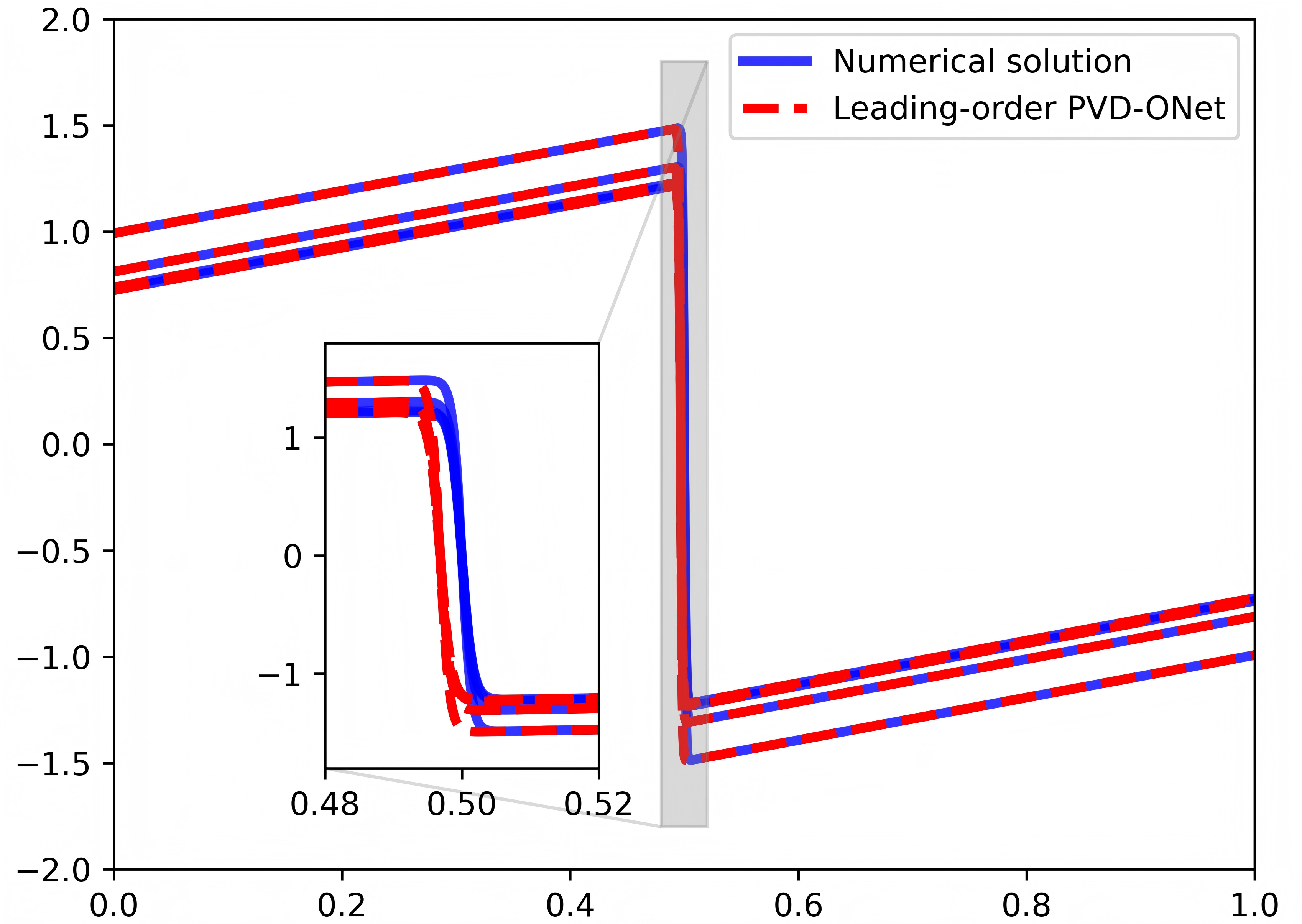}
    \end{subfigure}
    \hfill
    \begin{subfigure}[t]{0.44\linewidth}
        \centering
        \includegraphics[width=\linewidth]{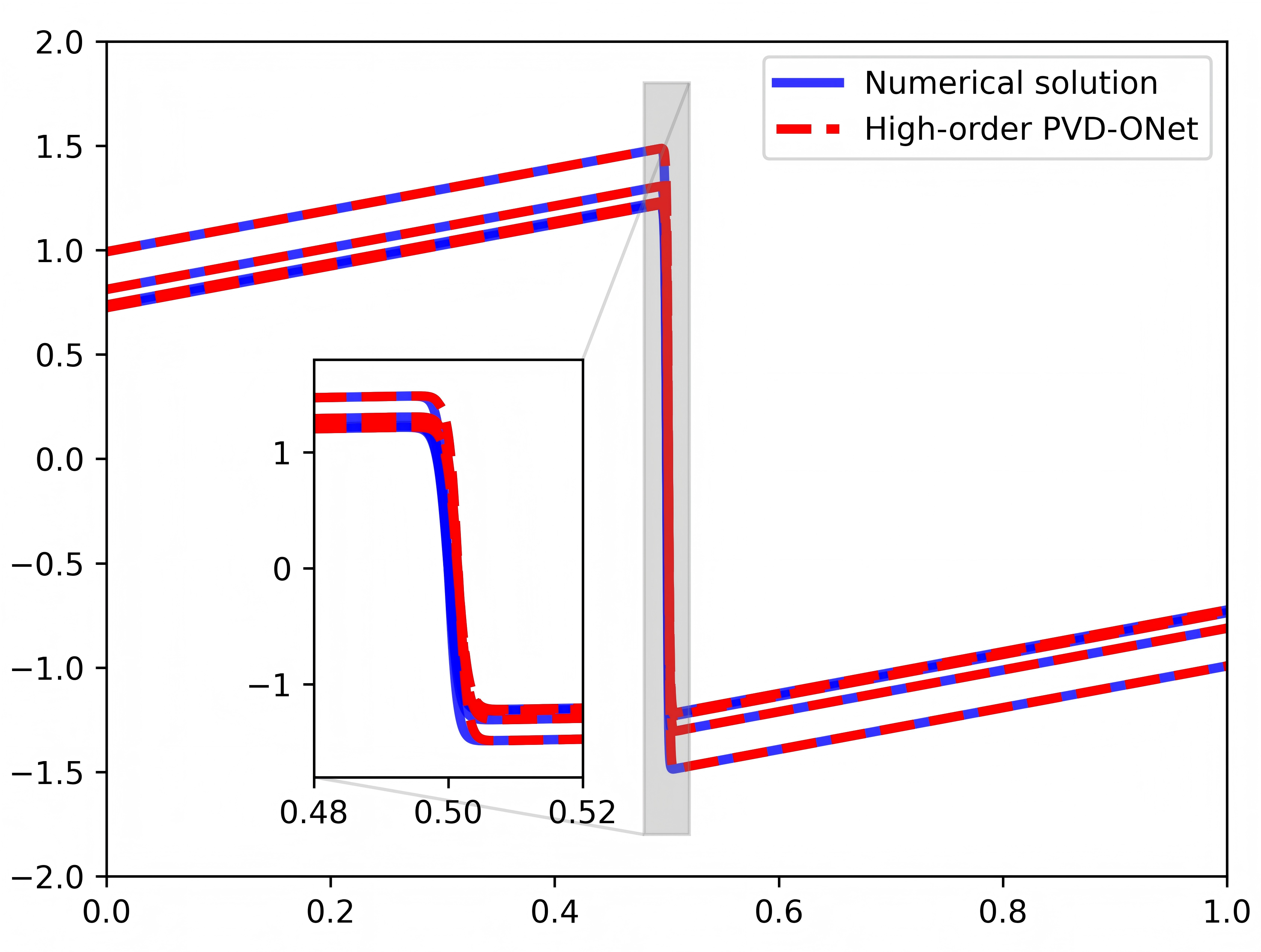}
    \end{subfigure}
    \caption{PVD-ONet for the internal layer problem. \textbf{(Left):} Predicted results of Leading-order PVD-ONet. \textbf{(Right):} Predicted results of High-order PVD-ONet. Our object is to learn the operator $G:(a,b)\mapsto u$. }
    \label{case3-PVD-ONet}
\end{figure}

\paragraph{PVD-ONet}
For the internal layer problem, we learn the operator $G:(a,b)\mapsto u$, where $a\sim \mathcal{U}[0.6,1]$ and $b=-a$. We use 500 boundary conditions for training and 100 for testing. As shown in Fig.~\ref{case3-PVD-ONet}, the leading-order PVD-ONet captures the overall solution trend but exhibits deviations near the internal layer, while the high-order version provides a more accurate approximation with reduced transition errors. Table~\ref{table:pvd-onet-internal} shows that PVD-ONet significantly outperforms PI-DeepONet, and the high-order PVD-ONet further improves accuracy over the leading-order version. 
\vspace{-25pt}

\begin{table}[H]
\centering
\caption{Comparison of mean prediction errors and training time (100,000 epochs) for PVD-ONet on the internal layer problem. Bold indicates the lowest error.}
\label{table:pvd-onet-internal}
\resizebox{\linewidth}{!}{%
\begin{tabular}{lccccc}
\toprule
\textbf{Method} 
& \multicolumn{2}{c}{\textbf{Global Errors}} 
& \multicolumn{2}{c}{\textbf{Inner Region Errors}} 
& \textbf{Training} \\
\cmidrule(r){2-3} \cmidrule(r){4-5}
& Relative $L^2$ & $L^\infty$ 
& Relative $L^2$ & $L^\infty$ 
& \textbf{Time (s)} \\
\midrule
PI-DeepONet \cite{wang2021learning}    
& $9.93 \times 10^{-1}$ & $1.32 \times 10^{0}$ 
& $9.95 \times 10^{-1}$ & $1.32 \times 10^{0}$ 
& 4,171 \\

Leading-order PVD-ONet                 
& $4.32 \times 10^{-1}$ & $2.01 \times 10^{0}$ 
& $4.35 \times 10^{-1}$ & $2.01 \times 10^{0}$ 
& 13,872 \\

\textbf{High-order PVD-ONet}            
& $\mathbf{2.05 \times 10^{-1}}$ & $\mathbf{1.10 \times 10^{0}}$ 
& $\mathbf{2.07 \times 10^{-1}}$ & $\mathbf{1.10 \times 10^{0}}$ 
& 47,031 \\
\bottomrule
\end{tabular}%
}
\end{table}

\subsection{2D coupled boundary layer problem}\label{2D-coupled}
To further investigate the applicability of the proposed PVD-Net and PVD-ONet, we consider a challenging and largely unexplored two-dimensional boundary layer problem featuring boundary layers in both the $x$- and $y$-directions and their interaction in a coupled corner layer.

\begin{figure}[h]
    \centering
    \includegraphics[width=0.8\linewidth]{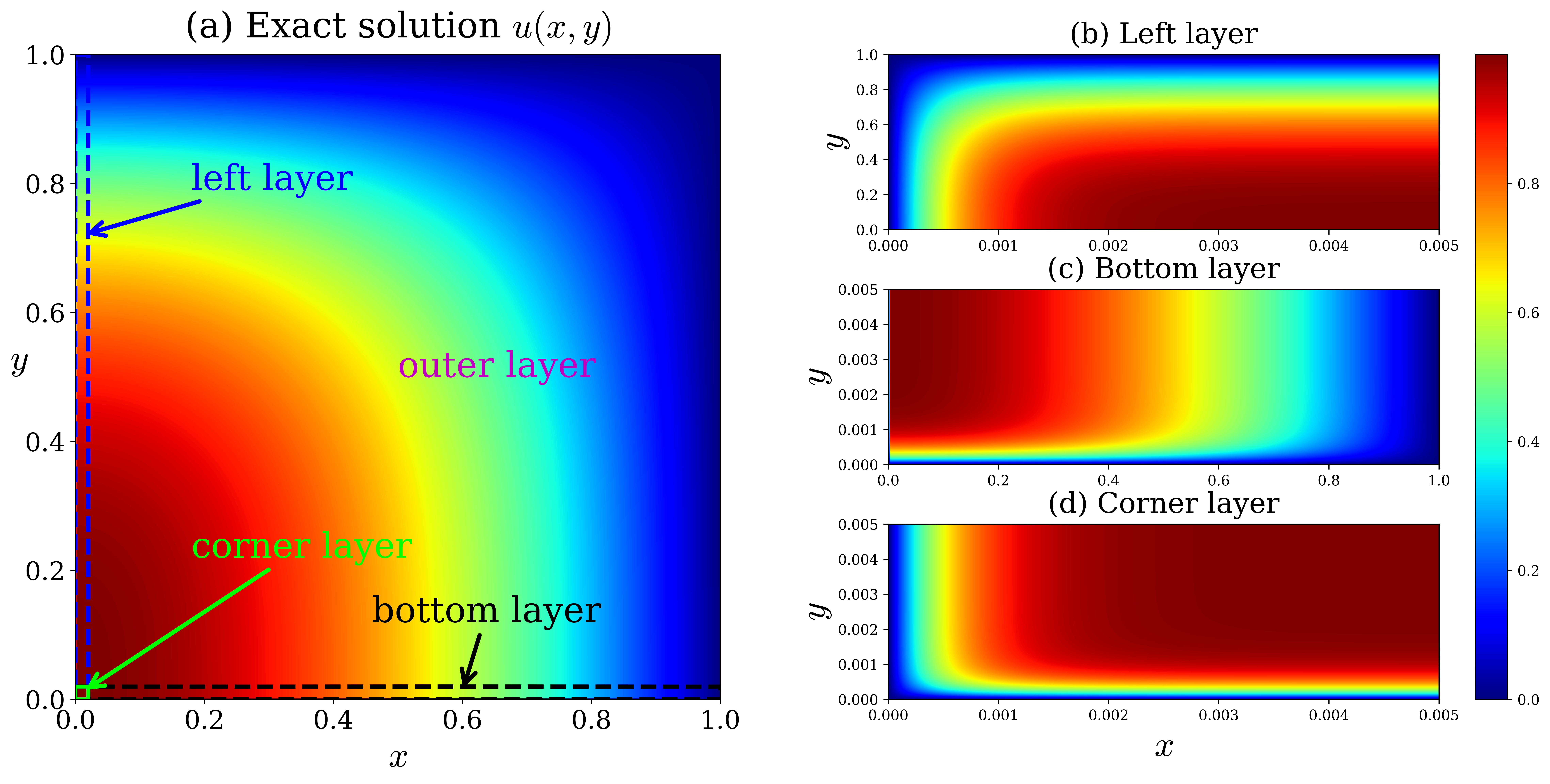}
    \caption{Exact solution of the 2D coupled boundary layer problem and localized zoomed-in views of the left boundary layer, bottom boundary layer, and corner layer.}
    \label{fig:2D_coupled_boundary_layer}
\end{figure}

Specifically, we consider the following two-dimensional convection--diffusion--reaction problem \cite{podila2024wavelet}:
\begin{align*}
\left\{
\begin{aligned}
-\varepsilon \Delta u
-(x+2)u_x
-(y^3+3)u_y
+u &= f(x,y),
&& (x,y)\in\Omega,\\
u &= 0,
&& (x,y)\in\partial\Omega,
\end{aligned}
\right.
\qquad
\Omega=(0,1)^2,
\end{align*}
where $\varepsilon=10^{-3}$.
The source term $f(x,y)$ is chosen such that the exact solution is
\begin{align}\label{case4-exact}
u(x,y)
=
\cos\left(\frac{\pi x}{2}\right)(1-y^3)
\left(1-e^{-2x/\varepsilon}\right)
\left(1-e^{-3y/\varepsilon}\right).
\end{align}
Unlike the one-dimensional examples considered previously, the present
problem contains two interacting boundary layers located at $x=0$ and
$y=0$, respectively. Their interaction additionally gives rise to a
two-dimensional corner layer near $(0,0)$, as illustrated in
Fig.~\ref{fig:2D_coupled_boundary_layer}. Therefore, two stretched
variables $\xi={x}/{\varepsilon}$,
$\eta={y}/{\varepsilon}$ are introduced 
to resolve the $x$- and $y$-boundary layers, respectively. We use the
same truncation parameter as in the previous examples,
$\xi_0=\eta_0=20$.

\paragraph{PVD-Net}
The leading-order PVD-Net consists of four neural networks: one outer network, two directional inner-layer networks, and one corner-layer network. The high-order PVD-Net employs nine networks, including two outer networks for the leading- and first-order approximations, three networks for each directional inner layer, and one corner-layer network. The corresponding asymptotic derivations are given in Section~SM6.3. In the loss function, the PDE residuals for the two boundary layers and the corner layer are weighted by 10, while all other terms use unit weight.
For evaluation, the $x$- and $y$-layer regions use $201\times51$ and $51\times201$ grids, respectively, and the corner region uses a $201\times201$ grid. Global errors are evaluated on a $353\times353$ grid over $[0,1]^2$. Along each coordinate direction, the 353 sampling points are formed by combining 101 and 51 uniformly spaced points over $[0,1]$ with 201 additional points densely sampled in $[0,0.01]$, providing finer resolution near the boundary layers.

\begin{figure}[!htbp]
    \centering
    \includegraphics[width=0.85\columnwidth]{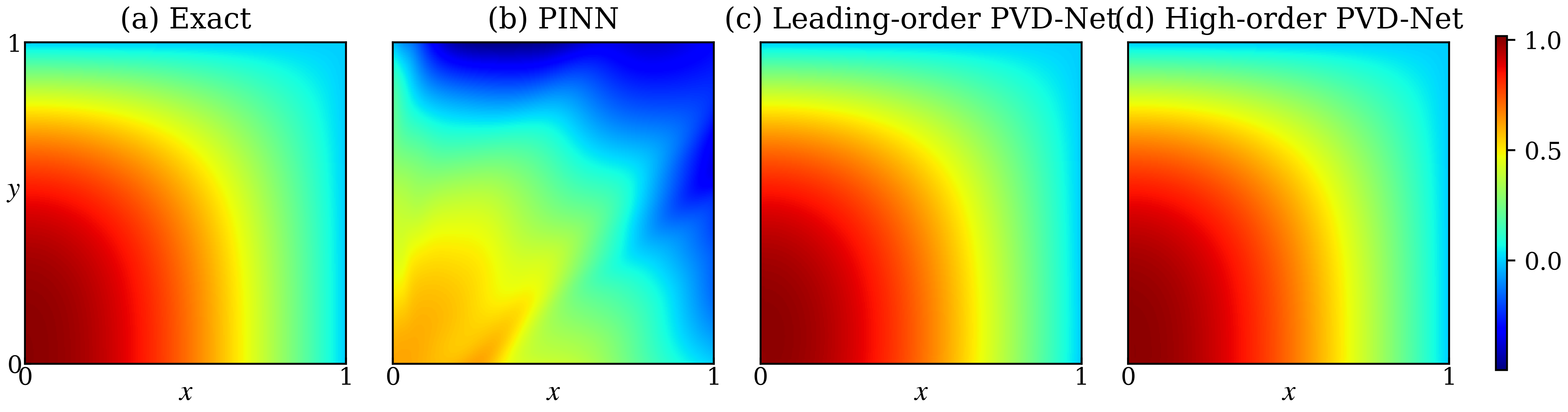}
    \caption{Global comparison of the exact solution, PINN, leading-order PVD-Net, and high-order PVD-Net.}
    \label{fig:global_comparison}
\end{figure}

\begin{figure}[!htbp]
    \centering
    \includegraphics[width=0.85\columnwidth]{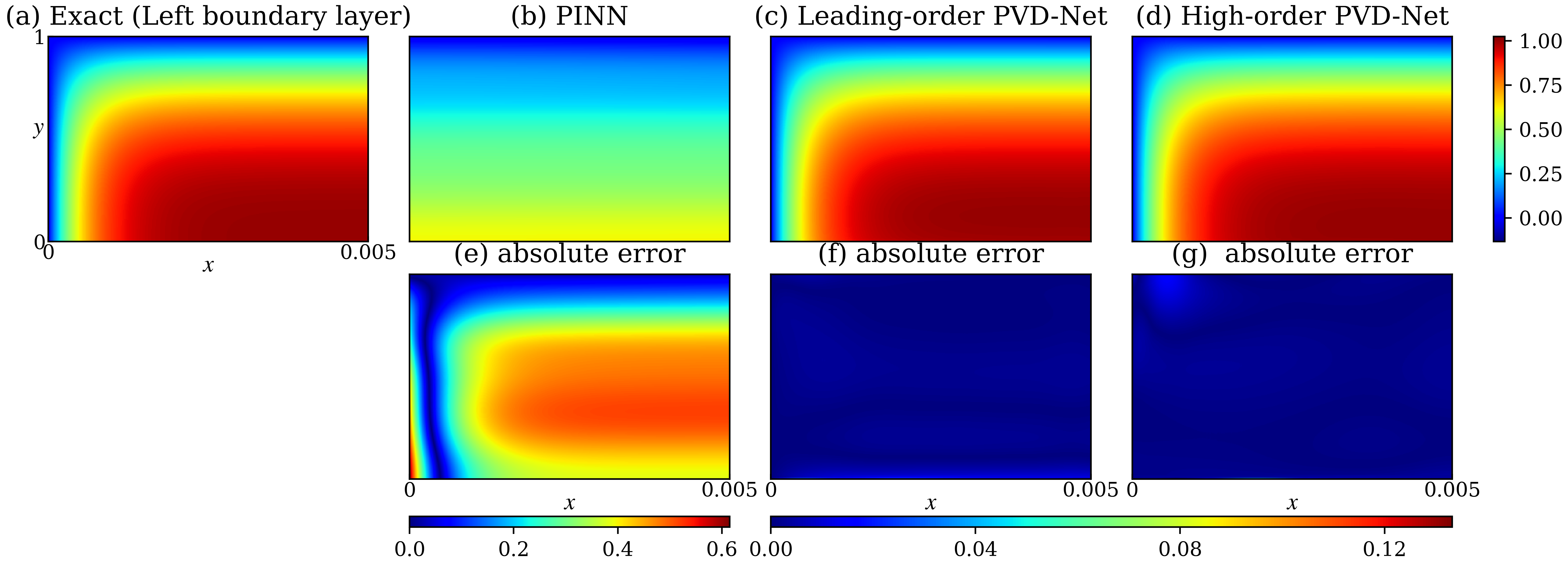}
    \caption{Predicted solutions and absolute errors in the left boundary-layer region.}
    \label{fig:left_layer}
\end{figure}

\begin{figure}[!htbp]
    \centering
    \includegraphics[width=0.85\columnwidth]{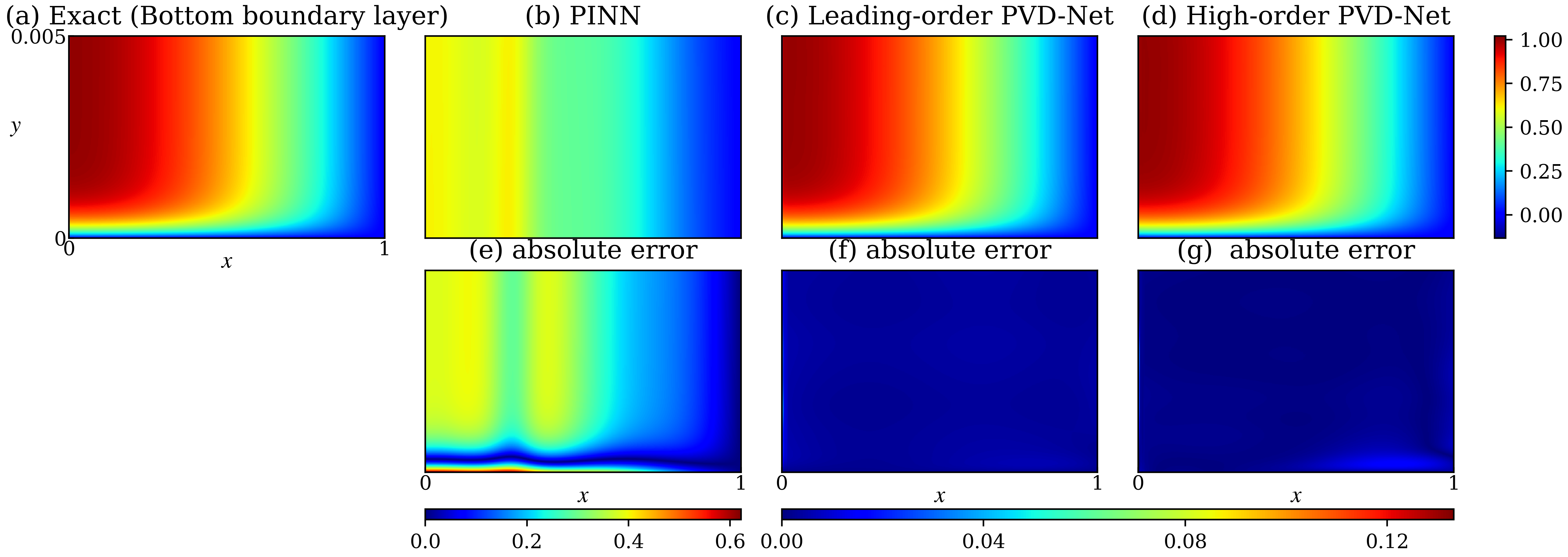}
    \caption{Predicted solutions and absolute errors in the bottom boundary-layer region.}
    \label{fig:bottom_layer}
\end{figure}

\begin{figure}[!htbp]
    \centering
    \includegraphics[width=0.85\columnwidth]{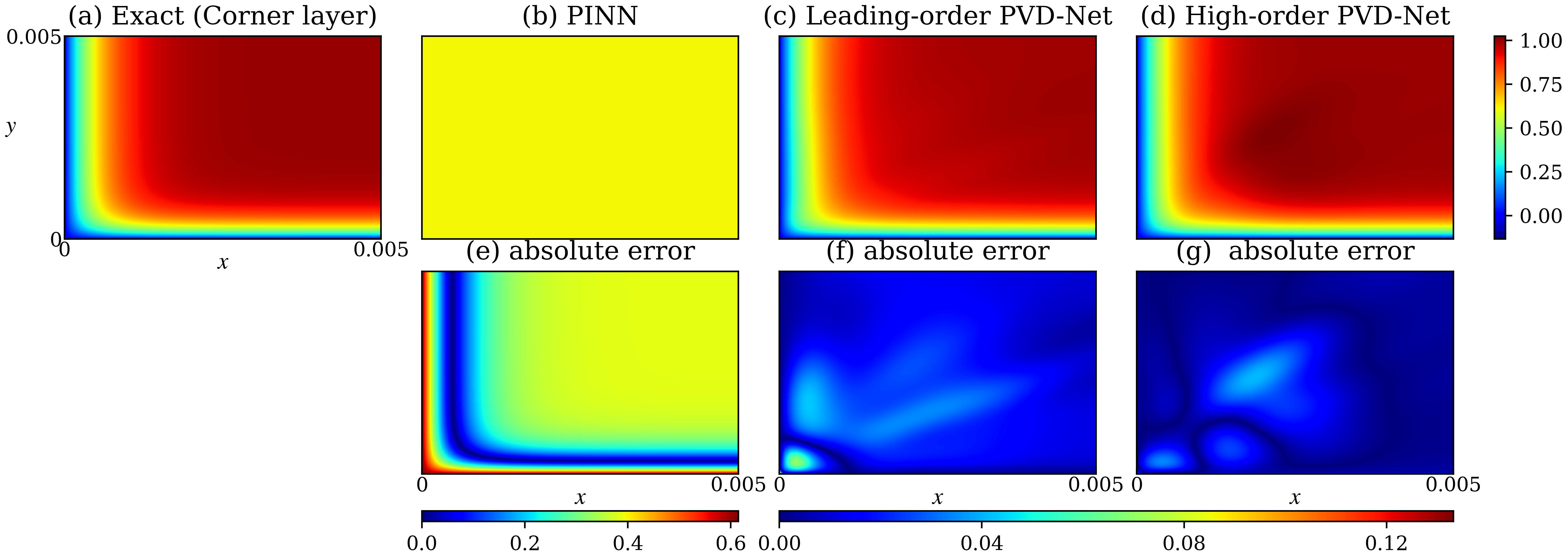}
    \caption{Predicted solutions and absolute errors in the in the corner-layer region.}
    \label{fig:corner_layer}
\end{figure}

Figure~\ref{fig:global_comparison} shows that PVD-Net accurately captures the solution structure, whereas the standard PINN fails to resolve the sharp boundary layers. The enlarged views in Figs.~\ref{fig:left_layer}--\ref{fig:corner_layer} further show that the high-order PVD-Net provides the closest agreement with the exact solution in the left, bottom, and corner layers. Table~\ref{tab:2d_comparison} confirms that it achieves lower relative $L^2$ and $L^\infty$ errors in all regions, with particularly pronounced improvements in the corner layer.

\begin{table}[h]
\centering
\caption{Comparison of prediction errors for the two-dimensional coupled boundary-layer problem after training for 100,000 epochs. The bolded values indicate the lowest errors among all methods.}
\label{tab:2d_comparison}
\setlength{\tabcolsep}{3.5pt}
\renewcommand{\arraystretch}{1.08}
\resizebox{\linewidth}{!}{
\begin{tabular}{lccccccccc}
\toprule
\textbf{Method}
& \multicolumn{2}{c}{\textbf{Global Errors}}
& \multicolumn{2}{c}{\textbf{$x$-Layer Errors}}
& \multicolumn{2}{c}{\textbf{$y$-Layer Errors}}
& \multicolumn{2}{c}{\textbf{Corner Errors}}
& \textbf{Training} \\
\cmidrule(r){2-3}
\cmidrule(r){4-5}
\cmidrule(r){6-7}
\cmidrule(r){8-9}
& Rel. $L^2$ & $L^\infty$
& Rel. $L^2$ & $L^\infty$
& Rel. $L^2$ & $L^\infty$
& Rel. $L^2$ & $L^\infty$
& \textbf{Time (s)} \\
\midrule
PINN
& $4.59\times10^{-1}$ & $6.21\times10^{-1}$
& $5.29\times10^{-1}$ & $6.15\times10^{-1}$
& $4.22\times10^{-1}$ & $6.21\times10^{-1}$
& $3.85\times10^{-1}$ & $6.15\times10^{-1}$
&  1,122\\

Leading-order PVD-Net
& $1.04\times10^{-2}$ & $1.33\times10^{-1}$
& $3.27\times10^{-3}$ & $1.33\times10^{-1}$
& $5.59\times10^{-3}$ & $1.33\times10^{-1}$
& $1.44\times10^{-2}$ & $1.33\times10^{-1}$
&2,864 \\

{High-order PVD-Net}
& $\mathbf{4.77\times10^{-3}}$ & $\mathbf{4.10\times10^{-2}}$
& $\mathbf{2.51\times10^{-3}}$ & $\mathbf{2.78\times10^{-2}}$
& $\mathbf{2.80\times10^{-3}}$ & $\mathbf{2.78\times10^{-2}}$
& $\mathbf{6.46\times10^{-3}}$ & $\mathbf{4.10\times10^{-2}}$
&11,792 \\
\bottomrule
\end{tabular}
}
\end{table}

\paragraph{PVD-ONet}
For the two-dimensional couple problem, we consider nonhomogeneous Dirichlet boundary conditions $u|_{\partial \Omega}=c$, where $c\sim\mathcal{U}[0,1]$, and learn the solution operator $G:c\mapsto u$. We use 1,000 boundary conditions for training and 100 for testing. The reference solutions are obtained using  finite difference method. Due to the increased computational cost of training multiple networks, we use 10,000 epochs for this example, which is sufficient to achieve convergence. The learning rate is set to $10^{-3}$. The loss weights for the boundary layer and corner PDE residuals are set to 10, while all other loss weights are set to 1. Figure~\ref{fig:pvd-onet-2d} presents the full-domain solutions and enlarged views of the corner layers for two test samples with $c=0.024$ and $c=0.824$. Both the leading-order and high-order PVD-ONet accurately reconstruct the solutions and capture the sharp boundary layer structures. The quantitative results in Table~\ref{tab:2d_pvd_onet_comparison} further demonstrate the superior overall accuracy of the high-order PVD-ONet.
\vspace{-20pt}
\begin{figure}[h]
    \centering
    \includegraphics[width=1\linewidth]{two_samples_corner_region.png}
    \caption{Comparison of full-domain solutions and corner layer regions for two test samples using PI-DeepONet, leading-order PVD-ONet, and high-order PVD-ONet.}
    \label{fig:pvd-onet-2d}
\end{figure}

\begin{table}[h]
\centering
\caption{Comparison of different methods for the two-dimensional coupled problem after 10,000 training epochs. Bold indicates the lowest error.}
\label{tab:2d_pvd_onet_comparison}
\setlength{\tabcolsep}{3.5pt}
\renewcommand{\arraystretch}{1.08}
\resizebox{\linewidth}{!}{
\begin{tabular}{lccccccccc}
\toprule
\textbf{Method}
& \multicolumn{2}{c}{\textbf{Global Errors}}
& \multicolumn{2}{c}{\textbf{$x$-Layer Errors}}
& \multicolumn{2}{c}{\textbf{$y$-Layer Errors}}
& \multicolumn{2}{c}{\textbf{Corner Errors}}
& \textbf{Training} \\
\cmidrule(r){2-3}
\cmidrule(r){4-5}
\cmidrule(r){6-7}
\cmidrule(r){8-9}
& Rel. $L^2$ & $L^\infty$
& Rel. $L^2$ & $L^\infty$
& Rel. $L^2$ & $L^\infty$
& Rel. $L^2$ & $L^\infty$
& \textbf{Time (s)} \\
\midrule
PI-DeepONet \cite{wang2021learning}
& $2.79\times10^{-1}$ & $6.05\times10^{-1}$
& $3.30\times10^{-1}$ & $5.76\times10^{-1}$
& $2.22\times10^{-1}$ & $6.05\times10^{-1}$
& $2.34\times10^{-1}$ & $5.76\times10^{-1}$
& 1,070 \\

Leading-order PVD-ONet
& $9.30\times10^{-3}$ & $6.29\times10^{-2}$
& $1.33\times10^{-2}$ & $5.22\times10^{-2}$
& $1.22\times10^{-2}$ & $6.29\times10^{-2}$
& $5.69\times10^{-3}$ & $3.79\times10^{-2}$
& 13,280 \\

High-order PVD-ONet
& $\mathbf{3.62\times10^{-3}}$ & $\mathbf{2.77\times10^{-2}}$
& $\mathbf{4.84\times10^{-3}}$ & $\mathbf{2.77\times10^{-2}}$
& $\mathbf{4.68\times10^{-3}}$ & $\mathbf{2.28\times10^{-2}}$
& $\mathbf{1.64\times10^{-3}}$ & $\mathbf{2.11\times10^{-2}}$
& 25,296 \\
\bottomrule
\end{tabular}
}
\end{table}

\subsection{Data-Driven Learning of the Scaling Exponent \texorpdfstring{$\lambda$}{lambda}}\label{inverse}

In the previous sections, the scaling exponent was fixed as $\lambda=1$, which assumes prior knowledge of the boundary layer scaling. Since the appropriate scaling is generally problem-dependent, we further treat $\lambda$ as a trainable parameter and infer it from a small number of data points sampled in the layer region. 
We consider three representative examples: a second-order equation with constant coefficients, a second-order equation with variable coefficients, and an internal layer problem. For all cases, the exact scaling exponent is $\lambda=1$, and only $N=10$ data points are used.
The learned scaling exponents for three examples are summarized in Table~\ref{tab:lambda}. In each case, the inferred value is close to the exact scaling $\lambda=1$, demonstrating that the boundary layer scaling can be recovered from only a small number of data points.
The detailed formulation, transformed equations, and loss functions are provided in Section SM7 of the Supplementary Material.

\begin{table}[h]
\centering
\caption{Learned scaling exponents for different examples using 10 data points. The exact value is $\lambda=1$.}
\label{tab:lambda}
\begin{tabular}{cc}
\toprule
\textbf{Example} & \textbf{Learned $\lambda$} \\
\midrule
Second-order equation with constant coefficients & 1.0100 \\
Second-order equation with variable coefficients & 1.0014 \\
Internal layer problem & 1.0131 \\
\bottomrule
\end{tabular}
\end{table}

\subsection{Asymptotic Convergence Rate Analysis}

\begin{figure}[h]
    \centering
    \includegraphics[width=0.9\linewidth]{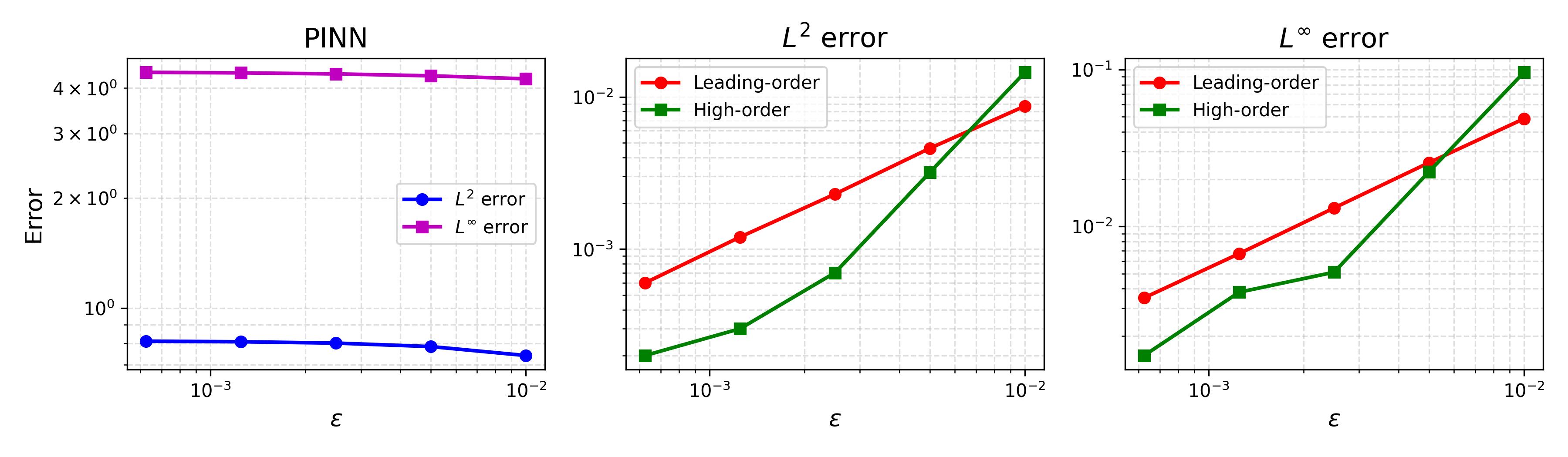}
    \caption{Log--log plots of the $L^2$ and $L^\infty$ errors versus $\varepsilon$. \textbf{(Left):} PINN shows no clear convergence. \textbf{(Middle and right):} The $L^2$ and $L^\infty$ errors of PVD-Net exhibit asymptotic convergence, with higher rates achieved by the high-order formulation.}
    \label{fig:log-log}
\end{figure}

To quantify the accuracy of the proposed framework, we investigate the asymptotic convergence behavior of the prediction error with respect to the perturbation parameter $\varepsilon$ using the constant-coefficient problem. Five representative values of $\varepsilon$ are considered: $0.000625$, $0.00125$, $0.0025$, $0.005$, and $0.01$. Figure~\ref{fig:log-log} presents the $L^2$ and $L^\infty$ errors versus $\varepsilon$ on a log--log scale. The approximately linear trends indicate a power-law decay of the form
\begin{equation*}
\|u_{\text{pred}} - u_{\text{true}}\| \sim \mathcal{O}(\varepsilon^p),
\end{equation*}
where $p$ denotes the convergence rate.

For the leading-order PVD-Net, the estimated convergence rates are $p \approx 0.97$ for the $L^2$ error and $p \approx 0.95$ for the $L^\infty$ error, consistent with the expected first-order accuracy. The high-order PVD-Net achieves improved rates of $p \approx 1.58$ and $p \approx 1.46$, respectively, demonstrating the effectiveness of higher-order asymptotic corrections. Although the high-order model can be less accurate for relatively large $\varepsilon$, it consistently outperforms the leading-order model as $\varepsilon$ becomes sufficiently small.

In contrast, the PINN baseline exhibits no clear asymptotic convergence, with both errors remaining nearly constant as $\varepsilon$ decreases. This behavior reflects the difficulty of resolving increasingly thin boundary layers due to spectral bias and the lack of explicit scale separation, highlighting the superior asymptotic convergence behavior of PVD-Net.

\section{Conclusion and Future Work}\label{conclusion}
In this work, we proposed two novel frameworks, PVD-Net and PVD-ONet, for solving boundary layer problems. Based on matched asymptotic expansions, we developed leading-order and high-order formulations using Prandtl and Van Dyke matching principles, respectively. The leading-order formulation employs two neural networks (or DeepONets) to construct uniformly valid composite solutions, emphasizing numerical stability. The high-order formulation incorporates five neural networks (or DeepONets) to resolve finer boundary layer structures and improve accuracy. Furthermore, the proposed framework extends to inverse problems by identifying the scaling exponent governing boundary layer thickness from sparse data. Numerical experiments on single boundary layer, internal layer, and two-dimensional coupled boundary layer problems demonstrate the effectiveness of the proposed methods.

For future research, we plan to explore several directions. First, adaptive mechanisms will be developed to automatically identify boundary layer locations based on problem-specific features, improving the flexibility and efficiency of the proposed frameworks. Second, rigorous theoretical analysis of the approximation errors and convergence properties of PVD-Net and PVD-ONet will be conducted to further understand their reliability and limitations.

\section*{Reproducibility of computational results} Additional mathematical derivations, implementation details, algorithms, and numerical results are available in the supplementary material.

\bibliographystyle{siamplain}
\bibliography{ref}
\end{document}